\documentclass{article}

\usepackage[final,nonatbib]{neurips_2018}

\usepackage[utf8]{inputenc} 
\usepackage[T1]{fontenc}    
\usepackage{hyperref}       
\usepackage{url}            
\usepackage{booktabs}       
\usepackage{amsfonts}       
\usepackage{nicefrac}       
\usepackage{microtype}      

\usepackage{amsmath,amssymb}       
\usepackage{bm}      
\usepackage[pdftex]{graphicx}
\usepackage{algorithm}
\usepackage[noend]{algpseudocode}
\usepackage{etoc}

\makeatletter
\def\algbackskip{\hskip-\ALG@thistlm}
\makeatother

\usepackage{color}

\newcommand{\ignore}[1]{}

\newcommand{\x}{\bm{x}}

\newcommand{\xx}{\text{\textbf{X}}}
\newcommand{\y}{\bm{y}}
\newcommand{\z}{\bm{z}}
\newcommand{\mmu}{\bm{\mu}}
\newcommand{\vmu}{\bm{\mu}}

\newcommand{\vlambda}{\bm{\lambda}}
\newcommand{\eps}{\bm{\varepsilon}}
\newcommand{\R}{{\text{\textbf{R}}}}
\newcommand{\vxi}{\bm{\xi}}

\newcommand{\X}{\mathcal{X}}

\newcommand{\nparams}{D}

\newcommand{\supplement}{Supplementary Material}

\newcommand{\normpdf}[3]{\mathcal{N}\left({#1}; {#2}, {#3 } \right)}

\newcommand{\data}{\mathcal{D}}

\newcommand{\like}{p(\data|\x)}
\newcommand{\post}{p(\x|\data)}
\newcommand{\ev}{p(\data)}

\newcommand{\qparams}{\bm{\phi}}

\newcommand{\qp}{q_{\qparams}}
\newcommand{\qx}{q(\x)}
\newcommand{\qpx}{q_{\qparams}(\x)}

\newcommand{\vgp}{\bm{\psi}}
\newcommand{\ngp}{n_\text{gp}}
\newcommand{\gpdata}{{\bm{\Xi}}}

\newcommand{\plb}{\text{PLB}}
\newcommand{\pub}{\text{PUB}}
\newcommand{\lb}{\text{LB}}
\newcommand{\ub}{\text{UB}}

\newcommand{\covar}{C}
\newcommand{\ns}{N_\text{s}}

\newcommand{\mat}[1]{\text{\bf{#1}}}
\newcommand{\ttheta}{\tilde{\theta}}

\newcommand{\noise}{\sigma_\text{obs}}

\algnewcommand\algorithmicinput{\textbf{Input:}}
\algnewcommand\INPUT{\item[\algorithmicinput]}

\algloop{Do}

\title{Variational Bayesian Monte Carlo}

\author{
Luigi Acerbi\thanks{Website: \url{luigiacerbi.com}. Alternative e-mail: \texttt{luigi.acerbi@gmail.com}.} 
\\
Department of Basic Neuroscience  \\
University of Geneva \\
\texttt{luigi.acerbi@unige.ch}
}

\begin{document}

\etocdepthtag.toc{mtchapter}
\etocsettagdepth{mtchapter}{subsubsection}
\etocsettagdepth{mtappendix}{none}

\maketitle
\setcounter{footnote}{0}

\vspace{-1em}
\begin{abstract}
\vspace{-0.25em}

Many probabilistic models of interest in scientific computing and machine learning have expensive, black-box likelihoods that prevent the application of standard techniques for Bayesian inference, such as MCMC, which would require access to the gradient or a large number of likelihood evaluations.
We introduce here a novel sample-efficient inference framework, Variational Bayesian Monte Carlo (VBMC). VBMC combines variational inference with Gaussian-process based, active-sampling Bayesian quadrature, using the latter to efficiently approximate the intractable integral in the variational objective.
Our method produces both a nonparametric approximation of the posterior distribution and an approximate lower bound of the model evidence, useful for model selection.
We demonstrate VBMC both on several synthetic likelihoods and on a neuronal model with data from real neurons. Across all tested problems and dimensions (up to $\nparams = 10$), VBMC performs consistently well in reconstructing the posterior and the model evidence with a limited budget of likelihood evaluations, unlike other methods that work only in very low dimensions. Our framework shows great promise as a novel tool for posterior and model inference with expensive, black-box likelihoods.
\end{abstract}

\vspace{-0.5em}
\section{Introduction}

\vspace{-0.25em}
In many scientific, engineering, and machine learning domains, such as in computational neuroscience and big data, complex black-box computational models are routinely used to estimate model parameters and compare hypotheses instantiated by different models.
Bayesian inference allows us to do so in a principled way that accounts for parameter and model uncertainty by computing the posterior distribution over parameters and the model evidence, also known as marginal likelihood or Bayes factor. However, Bayesian inference is generally analytically intractable, and the statistical tools of approximate inference, such as Markov Chain Monte Carlo (MCMC) or variational inference, generally require knowledge about the model (e.g., access to the gradients) and/or a large number of model evaluations. Both of these requirements cannot be met by black-box probabilistic models with computationally expensive likelihoods, precluding the application of standard Bayesian techniques of parameter and model uncertainty quantification to domains that would most need them.

Given a dataset $\data$ and model parameters $\x \in \mathbb{R}^\nparams$, here we consider the problem of computing both the \emph{posterior} $\post$ and the \emph{marginal likelihood} (or model evidence) 
$\ev$, defined as, respectively,
\begin{equation} \label{eq:postandml}
\post = \frac{\like p(\x)}{\ev} \qquad \text{and} \qquad \ev = \int \like p(\x) d\x,
\end{equation}
where $\like$ is the likelihood of the model of interest and $p(\x)$ is the prior over parameters. 
Crucially, we consider the case in which $\like$ is a black-box, expensive function for which we have a limited budget of function evaluations (of the order of few hundreds).

A promising approach to deal with such computational constraints consists of building a probabilistic model-based approximation of the function of interest, for example via Gaussian processes (GP) \cite{rasmussen2006gaussian}. This statistical surrogate can be used in lieu of the original, expensive function, allowing faster computations. Moreover, uncertainty in the surrogate can be used to actively guide sampling of the original function to obtain a better approximation in regions of interest for the application at hand.
This approach has been extremely successful in Bayesian optimization \cite{jones1998efficient,brochu2010tutorial,snoek2012practical,shahriari2016taking,acerbi2017practical} and in Bayesian quadrature for the computation of intractable integrals \cite{ohagan1991bayes,ghahramani2003bayesian}.

In particular, recent works have applied GP-based Bayesian quadrature to the estimation of the marginal likelihood \cite{ghahramani2003bayesian,osborne2012active,gunter2014sampling,briol2015frank}, and GP surrogates to build approximations of the posterior \cite{kandasamy2015bayesian,wang2017adaptive}. However, none of the existing approaches deals simultaneously with posterior and model inference. Moreover, it is unclear how these approximate methods would deal with likelihoods with realistic properties, such as medium dimensionality (up to $\nparams \sim 10$), mild multi-modality, heavy tails, and parameters that exhibit strong correlations---all common issues of real-world applications.

In this work, we introduce Variational Bayesian Monte Carlo (VBMC), a novel approximate inference framework that combines variational inference and active-sampling Bayesian quadrature via GP surrogates.\footnote{Code available at \url{https://github.com/lacerbi/vbmc}.} 
Our method affords simultaneous approximation of the posterior and of the model evidence in a sample-efficient manner. We demonstrate the robustness of our approach by testing VBMC and other inference algorithms on a variety of synthetic likelihoods with realistic, challenging properties. We also apply our method to a real problem in computational neuroscience, by fitting a model of neuronal selectivity in visual cortex \cite{goris2015origin}. Among the tested methods, VBMC is the only one with consistently good performance across problems, showing promise as a novel tool for posterior and model inference with expensive likelihoods in scientific computing and machine learning.

\vspace{-0.25em}
\section{Theoretical background}

\vspace{-0.25em}
\subsection{Variational inference}
\label{sec:variationalinference}

Variational Bayes is an approximate inference method whereby the posterior $\post$ is approximated by a simpler distribution $\qx \equiv \qpx$ that usually belongs to a parametric family \cite{jordan1999introduction,bishop2006pattern}. The goal of variational inference is to find the variational parameters $\qparams$ for which the variational posterior $\qp$ ``best'' approximates the true posterior. In variational methods, the mismatch between the two distributions is quantified by the Kullback-Leibler (KL) divergence,
\begin{equation} \label{eq:KL}
\text{KL}\left[\qpx || \post\right] = \mathbb{E}_{\qparams} \left[\log \frac{q_{\qparams}(\x)}{\post} \right],
\end{equation}
where we adopted the compact notation $\mathbb{E}_{\qparams} \equiv \mathbb{E}_{q_{\qparams}}$.
Inference is then reduced to an optimization problem, that is finding the variational parameter vector $\qparams$ that minimizes Eq. \ref{eq:KL}. 
We rewrite Eq. \ref{eq:KL} as
\begin{equation} \label{eq:refactoring}
\log \ev = \mathcal{F}[q_{\qparams}] + \text{KL}\left[\qpx || \post\right],
\end{equation}
where
\begin{equation} \label{eq:elbo}
\mathcal{F}\left[ q_{\qparams} \right] =  \mathbb{E}_{\qparams} \left[\log \frac{\like p(\x)}{q_{\qparams}(\x)} \right] = \mathbb{E}_{\qparams} \left[f(\x) \right] + \mathcal{H}[q_{\qparams}(\x)] 
\end{equation}
is the negative free energy, or \emph{evidence lower bound} (ELBO). Here $f(\x) \equiv \log \like p(\x) = \log p(\data, \x)$ is the log joint probability and $\mathcal{H}[q]$ is the entropy of $q$. Note that since the KL divergence is always non-negative, from Eq. \ref{eq:refactoring} we have $\mathcal{F}[q] \le \log \ev$, with equality holding if $\qx \equiv \post$. Thus, maximization of the variational objective, Eq. \ref{eq:elbo}, is equivalent to minimization of the KL divergence, and produces both an approximation of the posterior $\qp$ and a lower bound on the marginal likelihood, which can be used as a metric for model selection.

Normally, $q$ is chosen to belong to a family (e.g., a factorized posterior, or mean field) such that the expected log joint in Eq. \ref{eq:elbo} and the entropy can be computed analytically, possibly providing closed-form equations for a coordinate ascent algorithm. Here, we assume that $f(\x)$, like many computational models of interest, is an expensive black-box function, which prevents a direct computation of Eq. \ref{eq:elbo} analytically or via simple numerical integration.

\vspace{-0.25em}
\subsection{Bayesian quadrature}
\label{sec:bq}

Bayesian quadrature, also known as Bayesian Monte Carlo, is a means to obtain Bayesian estimates of the mean and variance of non-analytical integrals of the form $\langle f \rangle = \int f(\x) \pi(\x) d\x$, defined on a domain $\mathcal{X} = \mathbb{R}^\nparams$ \cite{ohagan1991bayes,ghahramani2003bayesian}. Here, $f$ is a function of interest and $\pi$ a known probability distribution. Typically, a Gaussian Process (GP) prior is specified for $f(\x)$.

\vspace{-0.5em}
\paragraph{Gaussian processes}

GPs are a flexible class of models for specifying prior distributions over unknown functions $f : \X \subseteq \mathbb{R}^{\nparams} \rightarrow \mathbb{R}$ \cite{rasmussen2006gaussian}. 
GPs are defined by a mean function $m: \X \rightarrow \mathbb{R}$  and a positive definite covariance, or kernel function $\kappa: \X \times \X \rightarrow \mathbb{R}$. In Bayesian quadrature, a common choice is the Gaussian kernel $\kappa(\x,\x^\prime) = \sigma_f^2 \normpdf{\x}{\x^\prime}{\bm{\Sigma}_\ell}$, with $\bm{\Sigma}_\ell = \text{diag}[{\ell^{(1)}}^2, \ldots, {\ell^{(\nparams)}}^2]$, where $\sigma_f$ is the output length scale and $\bm{\ell}$ is the vector of input length scales.
Conditioned on training inputs $\xx = \left\{\x_1, \ldots,\x_n \right\}$ and associated function values $\y = f(\xx)$, 
the GP posterior will have latent posterior conditional mean $\overline{f}_{\gpdata}(\x) \equiv \overline{f}(\x; \gpdata, \vgp)$ and covariance $\covar_\gpdata(\x, \x^\prime) \equiv \covar(\x, \x^\prime; \gpdata, \vgp)$ in closed form (see \cite{rasmussen2006gaussian}), where $\gpdata = \left\{\xx, \y \right\}$ is the set of training function data for the GP and $\vgp$ is a hyperparameter vector for the GP mean, covariance, and likelihood.

\vspace{-0.5em}
\paragraph{Bayesian integration}

Since integration is a linear operator, if $f$ is a GP, the posterior mean and variance of the integral $\int f(\x) \pi(\x) d\x$ are \cite{ghahramani2003bayesian}
\begin{equation} \label{eq:bq}
\mathbb{E}_{f | \gpdata}[\langle f \rangle] = \int \overline{f}_\gpdata(\x) \pi(\x) d\x, \qquad
\mathbb{V}_{f | \gpdata}[\langle f \rangle] = \int \int \covar_\gpdata(\x, \x^\prime) \pi(\x) d\x \pi(\x^\prime) d\x^\prime.
\end{equation}
Crucially, if $f$ has a Gaussian kernel and $\pi$ is a Gaussian or mixture of Gaussians (among other functional forms), the integrals in Eq. \ref{eq:bq} can be computed analytically. 

\vspace{-0.5em}
\paragraph{Active sampling}

For a given budget of samples $n_\text{max}$, a smart choice of the input samples $\xx$ would aim to minimize the posterior variance of the final integral (Eq. \ref{eq:bq}) \cite{briol2015frank}. Interestingly, for a standard GP and fixed GP hyperparameters $\vgp$, the optimal variance-minimizing design does not depend on the function values at $\xx$, thereby allowing precomputation of the optimal design. However, if the GP hyperparameters are updated online, or the GP is warped (e.g., via a log transform \cite{osborne2012active} or a square-root transform \cite{gunter2014sampling}), the variance of the posterior will depend on the function values obtained so far, and an active sampling strategy is desirable. The \emph{acquisition function} $a: \X \rightarrow \mathbb{R}$ determines which point in $\X$ should be evaluated next via a proxy optimization $\x_\text{next} = \text{argmax}_{\x} a(\x)$. Examples of acquisition functions for Bayesian quadrature include the \emph{expected entropy}, which minimizes the expected entropy of the integral after adding $\x$ to the training set \cite{osborne2012active}, and the much faster to compute \emph{uncertainty sampling} strategy, which maximizes the variance of the \emph{integrand} at $\x$ \cite{gunter2014sampling}.

\vspace{-0.25em}
\section{Variational Bayesian Monte Carlo (VBMC)}
\label{sec:bads}

\begin{figure}[htb]
  \includegraphics[width=\linewidth]{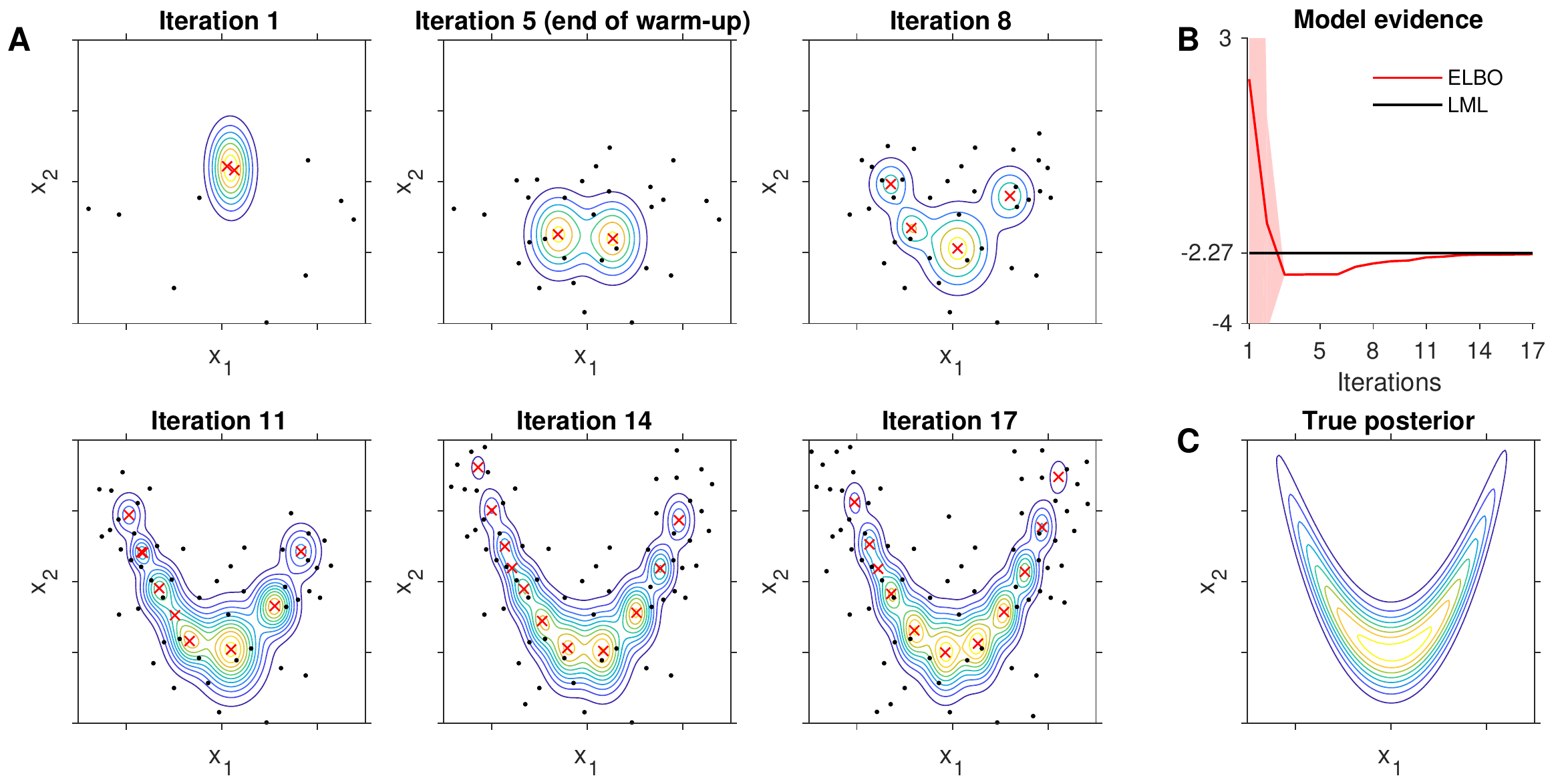}  
\vspace{-1.75em}
  \caption{{\bf Example run of VBMC on a 2-D pdf.} \textbf{A.} Contour plots of the variational posterior at different iterations of the algorithm. Red crosses indicate the centers of the variational mixture components, black dots are the training samples. \textbf{B.} ELBO as a function of iteration. Shaded area is 95\% CI of the ELBO in the current iteration as per the Bayesian quadrature approximation (\emph{not} the error wrt ground truth). The black line is the true log marginal likelihood (LML). \textbf{C.} True target pdf.
}
  \label{fig:demo}
\end{figure}

We introduce here Variational Bayesian Monte Carlo (VBMC), a sample-efficient inference method that combines variational Bayes and Bayesian quadrature, particularly useful for models with (moderately) expensive likelihoods. The main steps of VBMC are described in Algorithm \ref{alg:vbmc}, and an example run of VBMC on a nontrivial 2-D target density is shown in Fig. \ref{fig:demo}.

\vspace{-0.5em}
\paragraph {VBMC in a nutshell} 

In each iteration $t$, the algorithm: (1) sequentially samples a batch of `promising' new points that maximize a given acquisition function, and evaluates the (expensive) log joint $f$ at each of them; (2) trains a GP model of the log joint $f$, given the training set $\gpdata_t = \left\{\xx_t, \y_t \right\}$ of points evaluated so far; (3) updates the variational posterior approximation, indexed by $\qparams_t$, by optimizing the ELBO. This loop repeats until the budget of function evaluations is exhausted, or some other termination criterion is met (e.g., based on the stability of the found solution). VBMC includes an initial \emph{warm-up} stage to avoid spending computations in regions of low posterior probability mass (see Section \ref{subsec:warmup}). In the following sections, we describe various features of VBMC.

\begin{algorithm}[t]
\caption{Variational Bayesian Monte Carlo}\label{alg:vbmc}
\begin{algorithmic}[1]
\INPUT target log joint $f$, starting point $\bm{\x_0}$, plausible bounds \texttt{PLB}, \texttt{PUB}, additional \texttt{options}
\State \textbf{Initialization:} 
$t \leftarrow 0$, initialize variational posterior $\qparams_0$, \textsc{StopSampling} $\leftarrow$ \texttt{false} 
\Repeat
\State $t \leftarrow t + 1$
\If{$t \triangleq 1$}  \Comment{Initial design, Section \ref{subsec:warmup}}
\State Evaluate $y_0 \leftarrow f(\x_0)$ and add $(\x_0, y_0)$ to the training set $\gpdata$
\For{$2 \ldots n_\text{init}$}   
\State Sample a new point $\x_\text{new} \leftarrow \text{Uniform}[\texttt{PLB}, \texttt{PUB}]$
\State Evaluate $y_\text{new} \leftarrow f(\x_\text{new})$ and add $(\x_\text{new}, y_\text{new})$ to the training set $\gpdata$
\EndFor
\Else
\For{$1 \ldots n_\text{active}$}  \Comment{Active sampling, Section \ref{subsec:activesampling}} 
\State Actively sample a new point $\x_\text{new} \leftarrow \text{argmax}_{\x} a(\x)$
\State Evaluate $y_\text{new} \leftarrow f(\x_\text{new})$
and add $(\x_\text{new}, y_\text{new})$ to the training set $\gpdata$ \State \textbf{for each} $\vgp_1, \ldots, \vgp_{\ngp}$, perform rank-1 update of the GP posterior
\EndFor
\EndIf
\If{\textbf{not} \textsc{StopSampling}} \Comment{GP hyperparameter training, Section \ref{subsec:gphyp}}
\State $\{ \vgp_1, \ldots, \vgp_{\ngp} \} \leftarrow $ Sample GP hyperparameters 
\Else
\State $\vgp_1 \leftarrow $ Optimize GP hyperparameters
\EndIf
\State $K_t \leftarrow$ Update number of variational components \Comment{Section \ref{sec:adaptiveK}}
\State $\qparams_t \leftarrow$ Optimize ELBO via stochastic gradient descent \Comment{Section \ref{sec:elbo}}
\State Evaluate whether to \textsc{StopSampling} and other \textsc{TerminationCriteria}
\Until{\texttt{fevals} $>$ \texttt{MaxFunEvals} \textbf{ or } \textsc{TerminationCriteria}} \Comment{Stopping criteria, Section \ref{sec:stopping}}
\State \Return variational posterior $\qparams_t$, $\mathbb{E}\left[\text{ELBO}\right]$, $\sqrt{\mathbb{V}\left[\text{ELBO}\right]}$
\end{algorithmic}
\end{algorithm}

\vspace{-0.25em}
\subsection{Variational posterior}
\label{sec:vp}

We choose for the variational posterior $q(\x)$ a flexible ``nonparametric'' family, a mixture of $K$ Gaussians with shared covariances, modulo a scaling factor, 
\begin{equation} \label{eq:varpost}
q(\x) \equiv q_{\qparams}(\x) = \sum_{k = 1}^K w_k \normpdf{\x}{\mmu_k}{\sigma_k^2 \mathbf{\Sigma}},
\end{equation}
where $w_k$, $\mmu_k$, and $\sigma_k$ are, respectively, the mixture weight, mean, and scale of the $k$-th component, and $\mathbf{\Sigma}$  is a covariance matrix common to all elements of the mixture. In the following, we assume a diagonal matrix $\mathbf{\Sigma} \equiv \text{diag}[{{\lambda}^{(1)}}^2,\ldots,{\lambda^{(\nparams)}}^2]$.
The variational posterior for a given number of mixture components $K$ is parameterized by $\qparams \equiv (w_1,\ldots,w_K,\vmu_1, \ldots, \vmu_K, \sigma_1, \ldots, \sigma_K, \vlambda)$, which has $K(\nparams+2) + \nparams$ parameters. The number of components $K$ is set adaptively (see Section \ref{sec:adaptiveK}).

\vspace{-0.25em}
\subsection{The evidence lower bound}
\label{sec:elbo}

We approximate the ELBO (Eq. \ref{eq:elbo}) in two ways. First, we approximate the log joint probability $f$ with a GP with a squared exponential (rescaled Gaussian) kernel, a Gaussian likelihood with observation noise $\sigma_\text{obs} > 0$ (for numerical stability \cite{gramacy2012cases}), and a \emph{negative quadratic} mean function, defined as
\begin{equation}
m(\x) = m_0 - \frac{1}{2} \sum_{i=1}^{\nparams} \frac{\left(x^{(i)} - x_\text{m}^{(i)}\right)^2}{{\omega^{(i)}}^2},
\end{equation}
where $m_0$ is the maximum value of the mean, $\x_\text{m}$ is the location of the maximum, and $\bm{\omega}$ is a vector of length scales.
This mean function, unlike for example a constant mean, ensures that the posterior GP predictive mean $\overline{f}$ is a proper log probability distribution (that is, it is integrable when exponentiated).
Crucially, our choice of variational family (Eq. \ref{eq:varpost}) and kernel, likelihood and mean function of the GP affords an analytical computation of the posterior mean and variance of the expected log joint $\mathbb{E}_{\qparams} \left[f \right]$ (using Eq. \ref{eq:bq}), and of their gradients (see \supplement{} for details).
Second, we approximate the entropy of the variational posterior, $\mathcal{H}\left[ q_{\qparams}\right]$, via simple Monte Carlo sampling, and we propagate its gradient through the samples via the reparametrization trick \cite{kingma2013auto,miller2016variational}.\footnote{We also tried a deterministic approximation of the entropy proposed in \cite{gershman2012nonparametric}, with mixed results.}
Armed with expressions for the mean expected log joint, the entropy, and their gradients, we can efficiently optimize the (negative) mean ELBO via stochastic gradient descent \cite{kingma2014adam}.

\vspace{-0.5em}
\paragraph{Evidence lower confidence bound} We define the \emph{evidence lower confidence bound} (ELCBO) as
\begin{equation} \label{eq:elcbo}
\text{ELCBO}(\qparams,f) =  \mathbb{E}_{f | \gpdata}\left[\mathbb{E}_{\qparams} \left[f \right]\right] + \mathcal{H}[q_{\qparams}] - \beta_\text{LCB} \sqrt{\mathbb{V}_{f | \gpdata}\left[ \mathbb{E}_{\qparams} \left[f \right] \right]}
\end{equation}
where the first two terms are the ELBO (Eq. \ref{eq:elbo}) estimated via Bayesian quadrature, and the last term is the uncertainty in the computation of the expected log joint multiplied by a risk-sensitivity parameter $\beta_\text{LCB}$ (we set $\beta_\text{LCB} = 3$ unless specified otherwise). Eq. \ref{eq:elcbo} establishes a probabilistic lower bound on the ELBO, used to assess the improvement of the variational solution (see following sections).

\vspace{-0.25em}
\subsection{Active sampling}
\label{subsec:activesampling}

In VBMC, we are performing active sampling to compute a \emph{sequence} of integrals
$\mathbb{E}_{\qparams_1} \left[f \right], \mathbb{E}_{\qparams_2} \left[f \right], \ldots, \mathbb{E}_{\qparams_T} \left[f \right]$,
across iterations $1,\ldots, T$ such that (1) the sequence of variational parameters $\qparams_t$ converges to the variational posterior that minimizes the KL divergence with the true posterior, and (2) we have minimum variance on our final estimate of the ELBO. Note how this differs from active sampling in simple Bayesian quadrature, for which we only care about minimizing the variance of a single fixed integral. The ideal acquisition function for VBMC will correctly balance exploration of uncertain regions and exploitation of regions with high probability mass to ensure a fast convergence of the variational posterior as closely as possible to the ground truth.

We describe here two acquisition functions for VBMC based on uncertainty sampling. Let $V_{\gpdata}(\x) \equiv C_{\gpdata}(\x,\x)$ be the posterior GP variance at $\x$ given the current training set $\gpdata$. `Vanilla' uncertainty sampling for $\mathbb{E}_{\qparams} \left[f \right]$ is $a_\text{us}(\x) = V_{\gpdata}(\x) q_{\qparams}(\x)^2$, where $q_{\qparams}$ is the current variational posterior. Since $a_\text{us}$ only maximizes the variance of the integrand under the \emph{current} variational parameters, we expect it to be lacking in exploration. 
To promote exploration, we introduce \emph{prospective uncertainty sampling},
\begin{equation}
a_\text{pro}(\x) = V_{\gpdata}(\x) q_{\qparams}(\x) \exp\left(\overline{f}_{\gpdata}(\x)\right),
\end{equation}
where $\overline{f}_{\gpdata}$ is the GP posterior predictive mean. 
$a_\text{pro}$ aims at reducing uncertainty of the variational objective both for the current posterior and at prospective locations where the variational posterior might move to in the future, if not already there (high GP posterior mean).
The variational posterior in $a_\text{pro}$ acts as a regularizer,  preventing active sampling from following too eagerly fluctuations of the GP mean. For numerical stability of the GP, we include in all acquisition functions a regularization factor to prevent selection of points too close to existing training points (see \supplement{}).


At the beginning of each iteration after the first, VBMC actively samples $n_\text{active}$ points ($n_\text{active} = 5$ by default in this work). We select each point sequentially, by optimizing the chosen acquisition function via CMA-ES \cite{hansen2003reducing}, and apply fast rank-one updates of the GP posterior after each acquisition.

\vspace{-0.25em}
\subsection{Adaptive treatment of GP hyperparameters}
\label{subsec:gphyp}

The GP model in VBMC has $3\nparams+3$ hyperparameters, $\vgp = (\bm{\ell}, \sigma_f, \sigma_\text{obs}, m_0, \x_\text{m}, \bm{\omega})$.
We impose an empirical Bayes prior on the GP hyperparameters based on the current training set (see \supplement{}), 
and we sample from the posterior over hyperparameters via slice sampling \cite{neal2003slice}.
In each iteration, we collect $\ngp = \text{round}(80/\sqrt{n})$ samples, where $n$ is the size of the current GP training set, with the rationale that we require less samples as the posterior over hyperparameters becomes narrower due to more observations.
Given samples $\{\vgp\} \equiv \{ \vgp_1, \ldots, \vgp_{\ngp} \}$, and a random variable $\chi$ that depends on $\vgp$, we compute the expected mean and variance of $\chi$ as
\begin{equation} \label{eq:gpmarg}
\mathbb{E}\left[\chi | {\{\vgp\}}\right] = \frac{1}{\ngp} \sum_{j = 1}^{\ngp} \mathbb{E} \left[ \chi| \vgp_{j}\right], \quad \mathbb{V}\left[ \chi| {\{\vgp\}} \right] = \frac{1}{\ngp} \sum_{j = 1}^{\ngp} \mathbb{V} \left[ \chi| \vgp_{j} \right] + \text{Var}\left[ \left\{ \mathbb{E} \left[ \chi| \vgp_{j}\right] \right\}_{j=1}^{\ngp} \right],
\end{equation}
where $\text{Var}[\cdot]$ is the sample variance.  We use Eq. \ref{eq:gpmarg} to compute the GP posterior predictive mean and variances for the acquisition function, and to marginalize the expected log joint over hyperparameters.

The algorithm adaptively switches to a faster maximum-a-posteriori (MAP) estimation of the hyperparameters (via gradient-based optimization) when the additional variability of the expected log joint brought by multiple samples falls below a threshold for several iterations, 
a signal that sampling is bringing little advantage to the precision of the computation.

\vspace{-0.25em}
\subsection{Initialization and warm-up}
\label{subsec:warmup}

The algorithm is initialized by providing a starting point $\x_0$ (ideally, in a region of high posterior probability mass) and vectors of \emph{plausible} lower/upper bounds \texttt{PLB}, \texttt{PUB}, that identify a region of high posterior probability mass in parameter space. In the absence of other information, we obtained good results with plausible bounds containing the peak of prior mass in each coordinate dimension, such as the top $\sim 0.68$ probability region (that is, mean $\pm$ 1 SD for a Gaussian prior). 
The initial design consists of the provided starting point(s) $\x_0$ and additional points generated uniformly at random inside the plausible box, for a total of $n_\text{init} = 10$ points. The plausible box also sets the reference scale for each variable, and in future work might inform other aspects of the algorithm \cite{acerbi2017practical}.
The VBMC algorithm works in an unconstrained space ($\x \in \mathbb{R}^\nparams$), but bound constraints to the variables can  be easily handled
via a nonlinear remapping of the input space, with an appropriate Jacobian correction of the log probability density 
\cite{carpenter2017stan} (see Section \ref{subsec:goris2015} and \supplement{}).\footnote{The available code for VBMC currently supports both unbounded variables and bound constraints.}

\vspace{-0.5em}
\paragraph{Warm-up} We initialize the variational posterior with $K = 2$ components in the vicinity of $\x_0$, and with small values of $\sigma_1, \sigma_2$, and $\bm{\lambda}$ (relative to the width of the plausible box).
The algorithm starts in \emph{warm-up} mode, during which VBMC tries to quickly improve the ELBO by moving to regions with higher posterior probability. During warm-up, $K$ is clamped to only two components with $w_1 \equiv w_2 = 1/2$, and we collect a maximum of $\ngp = 8$ hyperparameter samples. 
Warm-up ends when the ELCBO (Eq. \ref{eq:elcbo}) shows an improvement of less than 1 for three consecutive iterations, suggesting that the variational solution has started to stabilize.
At the end of warm-up, we \emph{trim} the training set by removing points whose value of the log joint probability $y$ is more than $10 \cdot \nparams$ points lower than the maximum value $y_\text{max}$ observed so far. While not necessary in theory, we found that trimming generally increases the stability of the GP approximation, especially when VBMC is initialized in a region of very low probability under the true posterior. To allow the variational posterior to adapt, we do not actively sample new points in the first iteration after the end of warm-up.

\vspace{-0.25em}
\subsection{Adaptive number of variational mixture components}
\label{sec:adaptiveK}

After warm-up, we add and remove variational components following a simple set of rules. 

\vspace{-0.5em}
\paragraph{Adding components}

We define the current variational solution as \emph{improving} if the ELCBO of the last iteration is higher than the ELCBO in the past few iterations ($n_\text{recent} = 4$). In each iteration, we increment the number of components $K$ by 1 if the solution is improving and no mixture component was \emph{pruned} in the last iteration (see below). To speed up adaptation of the variational solution to a complex true posterior when the algorithm has nearly converged, we further add two extra components if the solution is \emph{stable} (see below) and no component was recently pruned. Each new component is created by splitting and jittering a randomly chosen existing component. We set a maximum number of components $K_\text{max} = n^{2/3}$, where $n$ is the size of the current training set $\gpdata$.

\vspace{-0.5em}
\paragraph{Removing components}

At the end of each variational optimization, we consider as a candidate for \emph{pruning} a random mixture component $k$ with mixture weight $w_k < w_\text{min}$. We recompute the ELCBO without the selected component (normalizing the remaining weights). If the `pruned' ELCBO differs from the original ELCBO less than $\varepsilon$, we remove the selected component. We iterate the process through all components with weights below threshold. For VBMC we set $w_\text{min} = 0.01$ and $\varepsilon = 0.01$.

\vspace{-0.25em}
\subsection{Termination criteria}
\label{sec:stopping}

At the end of each iteration, we assign a \emph{reliability index} $\rho(t) \ge 0$ to the current variational solution based on the following features: change in ELBO between the current and the previous iteration; estimated variance of the ELBO; KL divergence between the current and previous variational posterior (see \supplement{} for details). By construction, a $\rho(t) \lesssim 1$ is suggestive of a \emph{stable} solution. 
The algorithm terminates when obtaining a stable solution for $n_\text{stable} = 8$ iterations (with at most one non-stable iteration in-between), or when reaching a maximum number $n_\text{max}$ of function evaluations.
The algorithm returns the estimate of the mean and standard deviation of the ELBO (a lower bound on the marginal likelihood), and the variational posterior, from which we can cheaply draw samples for estimating distribution moments, marginals, and other properties of the posterior.
If the algorithm terminates before achieving long-term stability, it warns the user and returns a recent solution with the best ELCBO, using a conservative $\beta_\text{LCB} = 5$.

\vspace{-0.25em}
\section{Experiments}
\label{sec:experiments}

We tested VBMC and other common inference algorithms on several artificial and real problems consisting of a target likelihood and an associated prior. The goal of inference consists of approximating the posterior distribution and the log marginal likelihood (LML) with a fixed budget of likelihood evaluations, assumed to be (moderately) expensive.

\vspace{-0.5em}
\paragraph{Algorithms}

We tested VBMC with the `vanilla' uncertainty sampling acquisition function $a_\text{us}$ (VBMC-U) and with prospective uncertainty sampling, $a_\text{pro}$ (VBMC-P). We also tested simple Monte Carlo (SMC), annealed importance sampling (AIS), the original Bayesian Monte Carlo (BMC), doubly-Bayesian quadrature (BBQ  \cite{osborne2012active})\footnote{We also tested BBQ* (approximate GP hyperparameter marginalization), which perfomed similarly to BBQ.}, and warped sequential active Bayesian integration (WSABI, both in its linearized and moment-matching variants, WSABI-L and WSABI-M \cite{gunter2014sampling}).
For the basic setup of these methods, we follow \cite{gunter2014sampling}. Most of these algorithms only compute an approximation of the marginal likelihood based on a set of sampled points, but do not directly compute a posterior distribution. We obtain a posterior by training a GP model (equal to the one used by VBMC) on the log joint evaluated at the sampled points, and then drawing 2$\cdot 10^4$ MCMC samples from the GP posterior predictive mean via parallel slice sampling \cite{neal2003slice,gilks1994adaptive}. 
We also tested two methods for posterior estimation via GP surrogates, BAPE \cite{kandasamy2015bayesian} and AGP \cite{wang2017adaptive}. Since these methods only compute an approximate posterior, we obtain a crude estimate of the log normalization constant (the LML) as the average difference between the log of the approximate posterior and the evaluated log joint at the top 20\% points in terms of posterior density.
For all algorithms, we use default settings, allowing only changes based on knowledge of the mean and (diagonal) covariance of the provided prior.

\vspace{-0.5em}
\paragraph{Procedure}

For each problem, we allow a fixed budget of $50 \times (\nparams + 2)$ likelihood evaluations, where $\nparams$ is the number of variables.
Given the limited number of samples, we judge the quality of the posterior approximation in terms of its first two moments, by computing the ``Gaussianized'' symmetrized KL divergence (gsKL) between posterior approximation and ground truth. The gsKL is defined as the symmetrized KL between two multivariate normal distributions with mean and covariances equal, respectively, to the moments of the approximate posterior and the moments of the true posterior. We measure the quality of the approximation of the LML in terms of \emph{absolute} error from ground truth, the rationale being that differences of LML are used for model comparison. Ideally, we want the LML error to be of order 1 of less, since much larger errors could severely affect the results of a comparison (e.g., differences of LML of 10 points or more are often presented as \emph{decisive} evidence in favor of one model \cite{kass1995bayes}). On the other hand, errors $\lesssim0.1$ can be considered negligible; higher precision is unnecessary. For each algorithm, we ran at least 20 separate runs per test problem with different random seeds, and report the median gsKL and LML error and the 95\% CI of the median calculated by bootstrap.
For each run, we draw the starting point $\x_0$ (if requested by the algorithm) uniformly from a box within 1 prior standard deviation (SD) from the prior mean. We use the same box to define the plausible bounds for VBMC.

\vspace{-0.25em}
\subsection{Synthetic likelihoods}

\vspace{-0.5em}
\paragraph{Problem set}

We built a benchmark set of synthetic likelihoods belonging to three families that represent typical features of target densities (see \supplement{} for details). Likelihoods in the \emph{lumpy} family are built out of a mixture of 12 multivariate normals with component means drawn randomly in the unit $\nparams$-hypercube, distinct diagonal covariances with SDs in the $[0.2,0.6]$ range, and mixture weights drawn from a Dirichlet distribution with unit concentration parameter. The lumpy distributions are mildly multimodal, in that modes are nearby and connected by regions with non-neglibile probability mass. In the \emph{Student} family, the likelihood is a multivariate Student's $t$-distribution with diagonal covariance and degrees of freedom equally spaced in the $[2.5,2+D/2]$ range across different coordinate dimensions. These distributions have heavy tails which might be problematic for some methods. Finally, in the \emph{cigar} family the likelihood is a multivariate normal in which one axis is 100 times longer than the others, and the covariance matrix is non-diagonal after a random rotation. The cigar family tests the ability of an algorithm to explore non axis-aligned directions.
For each family, we generated test functions for $\nparams \in \{ 2, 4, 6, 8, 10\}$, for a total of 15 synthetic problems.
For each problem, we pick as a broad prior a multivariate normal with mean centered at the expected mean of the family of distributions, and diagonal covariance matrix with SD equal to 3-4 times the SD in each dimension. For all problems, we compute ground truth values for the LML and the posterior mean and covariance analytically or via multiple 1-D numerical integrals.

\vspace{-0.5em}
\paragraph{Results}

\begin{figure}[htb]
  \includegraphics[width=\linewidth]{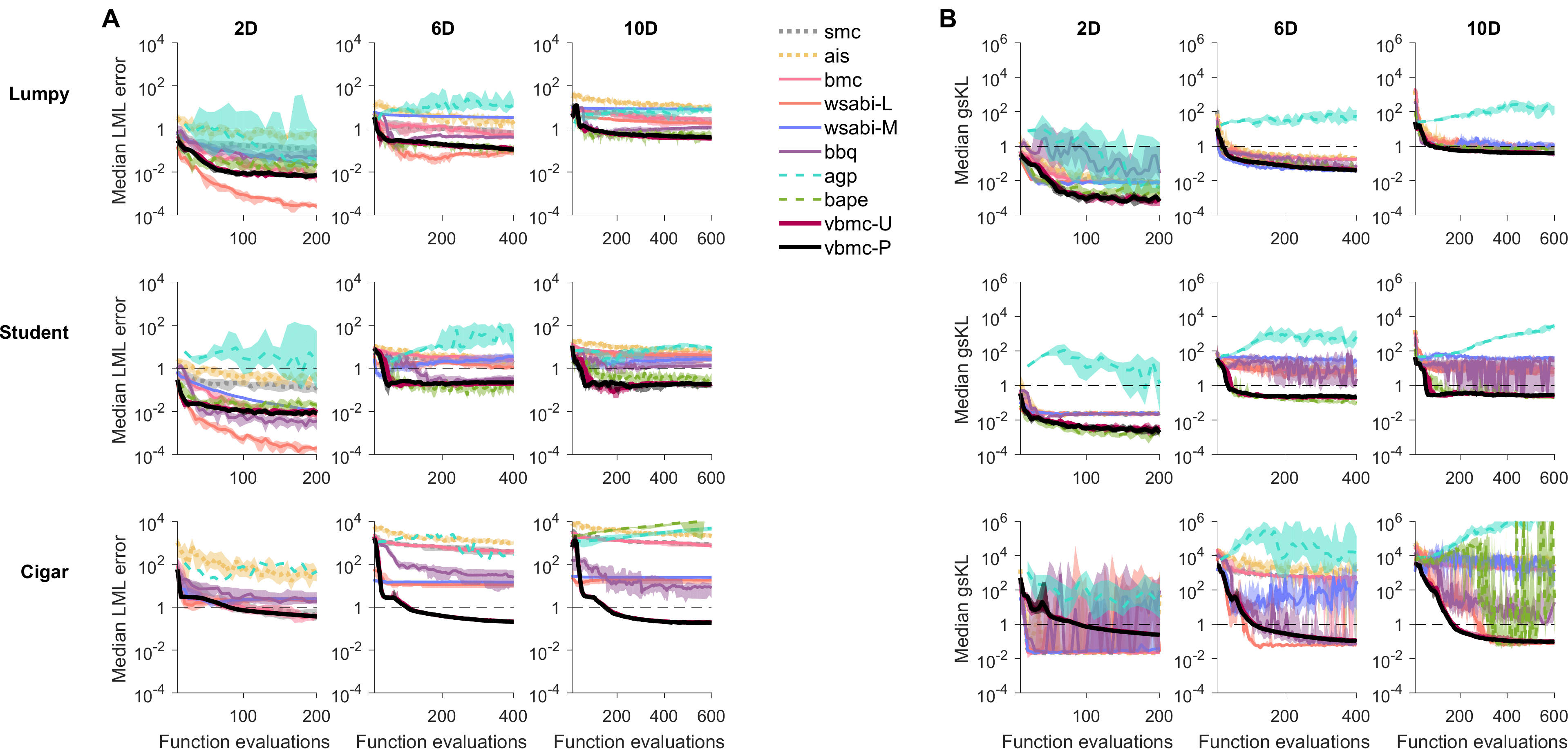}  
\vspace{-1.65em}
  \caption{{\bf Synthetic likelihoods.} \textbf{A.} Median absolute error of the LML estimate with respect to ground truth, as a function of number of likelihood evaluations, on the \emph{lumpy} (top), \emph{Student} (middle), and \emph{cigar} (bottom) problems, for $\nparams \in \{2, 6, 10\}$ (columns). \textbf{B.} Median ``Gaussianized'' symmetrized KL divergence between the algorithm's posterior and ground truth.
For both metrics, shaded areas are 95 \% CI of the median,
and we consider a desirable threshold to be below one (dashed line).
}
  \label{fig:synthetic}
\end{figure}

We show the results for $\nparams \in \{2,6,10\}$ in Fig. \ref{fig:synthetic} (see \supplement{} for full results, in higher resolution). Almost all algorithms perform reasonably well in very low dimension ($\nparams = 2$), and in fact several algorithms converge faster than VBMC to the ground truth (e.g., WSABI-L). However, as we increase in dimension, we see that all algorithms start failing, with only VBMC peforming consistently well across problems. In particular, besides the simple $\nparams = 2$ case, only VBMC obtains acceptable results for the LML with non-axis aligned distributions (\emph{cigar}). Some algorithms (such as AGP and BAPE) exhibited large numerical instabilities on the \emph{cigar} family, despite our best attempts at regularization, such that many runs were unable to complete.

\vspace{-0.25em}
\subsection{Real likelihoods of neuronal model}
\label{subsec:goris2015}

\vspace{-0.5em}
\paragraph{Problem set}

For a test with real models and data, we consider a computational model of neuronal orientation selectivity in visual cortex \cite{goris2015origin}. We fit the neural recordings of one V1 and one V2 cell with the authors' neuronal model that combines effects of filtering, suppression, and response nonlinearity \cite{goris2015origin}. 
The model is analytical but still computationally expensive due to large datasets and a cascade of several nonlinear operations.
For the purpose of our benchmark, we fix some parameters of the original model to their MAP values, yielding an inference problem with $\nparams = 7$ free parameters of experimental interest. We transform bounded parameters to uncontrained space via a logit transform \cite{carpenter2017stan}, and we place a broad Gaussian prior on each of the transformed variables, based on estimates from other neurons in the same study \cite{goris2015origin} (see \supplement{} for more details on the setup).
For both datasets, we computed the ground truth with $4 \cdot 10^5$ samples from the posterior, obtained via parallel slice sampling after a long burn-in. We calculated the ground truth LML from posterior MCMC samples via Geyer's reverse logistic regression \cite{geyer1994estimating}, and we independently validated it with a Laplace approximation, obtained via numerical calculation of the Hessian at the MAP (for both datasets, Geyer's and Laplace's estimates of the LML are within $\sim$ 1 point).

\begin{figure}[htb]
  \includegraphics[width=\linewidth]{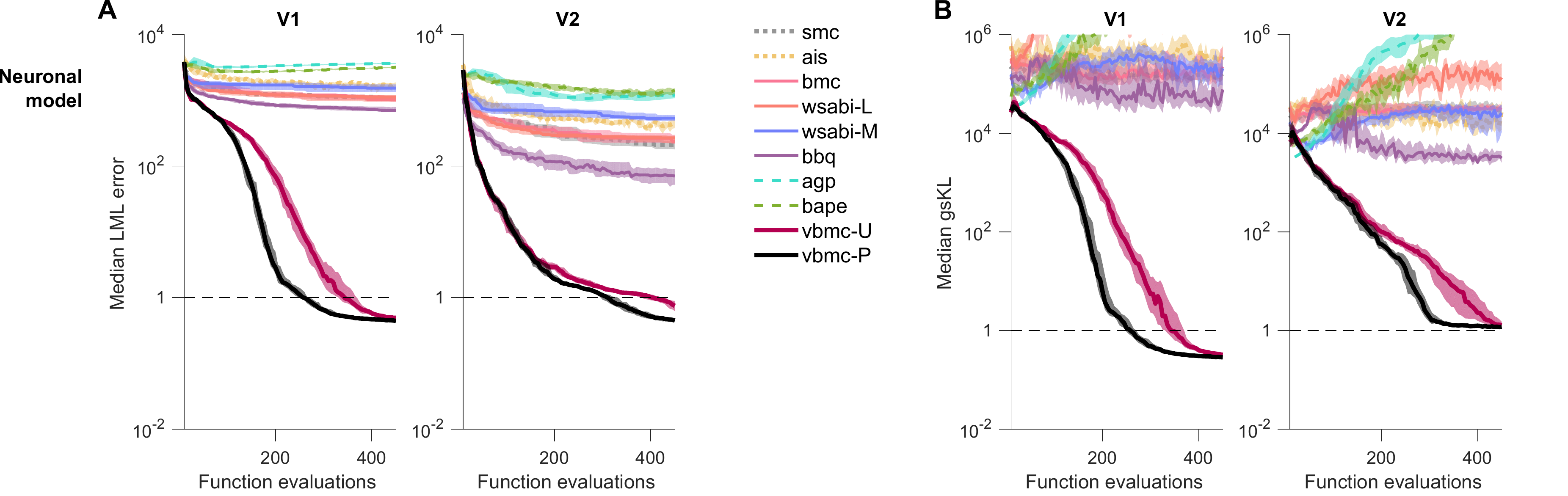}  
\vspace{-1.75em}
  \caption{{\bf Neuronal model likelihoods.} \textbf{A.} Median absolute error of the LML estimate, as a function of number of likelihood evaluations, for two distinct neurons ($\nparams = 7$). \textbf{B.} Median ``Gaussianized'' symmetrized KL divergence between the algorithm's posterior and ground truth. See also Fig. \ref{fig:synthetic}. 
}
  \label{fig:goris2015}
\end{figure}

\vspace{-0.5em}
\paragraph{Results}

For both datasets, VBMC is able to find a reasonable approximation of the LML and of the posterior, whereas no other algorithm produces a usable solution (Fig. \ref{fig:goris2015}). Importantly, the behavior of VBMC is fairly consistent across runs (see \supplement{}). We argue that the superior results of VBMC stem from a better exploration of the posterior landscape, and from a better approximation of the log joint (used in the ELBO), related but distinct features. To show this, we first trained GPs (as we did for the other methods) on the samples collected by VBMC (see \supplement{}). The posteriors obtained by sampling from the GPs trained on the VBMC samples scored a better gsKL than the other methods (and occasionally better than VBMC itself). Second, we estimated the marginal likelihood with WSABI-L using the samples collected by VBMC. The LML error in this hybrid approach is much lower than the error of WSABI-L alone, but still higher than the LML error of VBMC. These results combined suggest that VBMC builds better (and more stable) surrogate models and obtains higher-quality samples than the compared methods.

The performance of VBMC-U and VBMC-P is similar on synthetic functions, but the `prospective' acquisition function converges faster on the real problem set, so we recommend  $a_\text{pro}$ as the default.
Besides scoring well on quantitative metrics, VBMC is able to capture nontrivial features of the true posteriors (see \supplement{} for examples). Moreover, VBMC achieves these results with a relatively small computational cost (see \supplement{} for discussion).

\vspace{-0.25em}
\section{Conclusions}

In this paper, we have introduced VBMC, a novel Bayesian inference framework that combines variational inference with active-sampling Bayesian quadrature for models with expensive black-box likelihoods. Our method affords both posterior estimation and model inference by providing an approximate posterior and a lower bound to the model evidence. We have shown on both synthetic and real model-fitting problems that, given a contained budget of likelihood evaluations, VBMC is able to reliably compute valid, usable approximations in realistic scenarios, unlike previous methods whose applicability seems to be limited to very low dimension or simple likelihoods. Our method, thus, represents a novel useful tool for approximate inference in science and engineering.

We believe this is only the starting point to harness the combined power of variational inference and Bayesian quadrature. Not unlike the related field of Bayesian optimization, VBMC paves the way to a plenitude of both theoretical (e.g., analysis of convergence, development of principled acquisition functions) and applied work (e.g., application to case studies of interest, extension to noisy likelihood evaluations, algorithmic improvements), which we plan to pursue as future directions.

\clearpage

\subsection*{Acknowledgments}

We thank Robbe Goris for sharing data and code for the neuronal model; Michael Schartner and Rex Liu for comments on an earlier version of the paper; and three anonymous reviewers for useful feedback.

{\def\section*#1{}
{ 

\subsubsection*{References} 
\small

\begin{thebibliography}{10}

\bibitem{rasmussen2006gaussian}
Rasmussen, C. \& Williams, C. K.~I.
\newblock (2006) {\em {G}aussian {P}rocesses for {M}achine {L}earning}.
\newblock (MIT Press).

\bibitem{jones1998efficient}
Jones, D.~R., Schonlau, M.,  \& Welch, W.~J.
\newblock (1998) Efficient global optimization of expensive black-box
  functions.
\newblock {\em Journal of Global Optimization} {\bf 13}, 455--492.

\bibitem{brochu2010tutorial}
Brochu, E., Cora, V.~M.,  \& De~Freitas, N.
\newblock (2010) A tutorial on {B}ayesian optimization of expensive cost
  functions, with application to active user modeling and hierarchical
  reinforcement learning.
\newblock {\em arXiv preprint arXiv:1012.2599}.

\bibitem{snoek2012practical}
Snoek, J., Larochelle, H.,  \& Adams, R.~P.
\newblock (2012) Practical {B}ayesian optimization of machine learning
  algorithms.
\newblock {\em Advances in Neural Information Processing Systems} {\bf 25},
  2951--2959.

\bibitem{shahriari2016taking}
Shahriari, B., Swersky, K., Wang, Z., Adams, R.~P.,  \& de~Freitas, N.
\newblock (2016) Taking the human out of the loop: {A} review of {B}ayesian
  optimization.
\newblock {\em Proceedings of the IEEE} {\bf 104}, 148--175.

\bibitem{acerbi2017practical}
Acerbi, L. \& Ma, W.~J.
\newblock (2017) Practical {B}ayesian optimization for model fitting with
  {B}ayesian adaptive direct search.
\newblock {\em Advances in Neural Information Processing Systems} {\bf 30},
  1834--1844.

\bibitem{ohagan1991bayes}
O'Hagan, A.
\newblock (1991) {B}ayes--{H}ermite quadrature.
\newblock {\em Journal of Statistical Planning and Inference} {\bf 29},
  245--260.

\bibitem{ghahramani2003bayesian}
Ghahramani, Z. \& Rasmussen, C.~E.
\newblock (2002) {B}ayesian {M}onte {C}arlo.
\newblock {\em Advances in Neural Information Processing Systems} {\bf 15},
  505--512.

\bibitem{osborne2012active}
Osborne, M., Duvenaud, D.~K., Garnett, R., Rasmussen, C.~E., Roberts, S.~J.,
  \& Ghahramani, Z.
\newblock (2012) Active learning of model evidence using {B}ayesian quadrature.
\newblock {\em Advances in Neural Information Processing Systems} {\bf 25},
  46--54.

\bibitem{gunter2014sampling}
Gunter, T., Osborne, M.~A., Garnett, R., Hennig, P.,  \& Roberts, S.~J.
\newblock (2014) Sampling for inference in probabilistic models with fast
  {B}ayesian quadrature.
\newblock {\em Advances in Neural Information Processing Systems} {\bf 27},
  2789--2797.

\bibitem{briol2015frank}
Briol, F.-X., Oates, C., Girolami, M.,  \& Osborne, M.~A.
\newblock (2015) {F}rank-{W}olfe {B}ayesian quadrature: {P}robabilistic
  integration with theoretical guarantees.
\newblock {\em Advances in Neural Information Processing Systems} {\bf 28},
  1162--1170.

\bibitem{kandasamy2015bayesian}
Kandasamy, K., Schneider, J.,  \& P{\'o}czos, B.
\newblock (2015) {B}ayesian active learning for posterior estimation.
\newblock {\em Twenty-Fourth International Joint Conference on Artificial
  Intelligence}.

\bibitem{wang2017adaptive}
Wang, H. \& Li, J.
\newblock (2018) Adaptive {G}aussian process approximation for {B}ayesian
  inference with expensive likelihood functions.
\newblock {\em Neural Computation} pp. 1--23.

\bibitem{goris2015origin}
Goris, R.~L., Simoncelli, E.~P.,  \& Movshon, J.~A.
\newblock (2015) Origin and function of tuning diversity in macaque visual
  cortex.
\newblock {\em Neuron} {\bf 88}, 819--831.

\bibitem{jordan1999introduction}
Jordan, M.~I., Ghahramani, Z., Jaakkola, T.~S.,  \& Saul, L.~K.
\newblock (1999) An introduction to variational methods for graphical models.
\newblock {\em Machine Learning} {\bf 37}, 183--233.

\bibitem{bishop2006pattern}
Bishop, C.~M.
\newblock (2006) {\em Pattern Recognition and Machine Learning}.
\newblock (Springer).

\bibitem{gramacy2012cases}
Gramacy, R.~B. \& Lee, H.~K.
\newblock (2012) Cases for the nugget in modeling computer experiments.
\newblock {\em Statistics and Computing} {\bf 22}, 713--722.

\bibitem{kingma2013auto}
Kingma, D.~P. \& Welling, M.
\newblock (2013) Auto-encoding variational {B}ayes.
\newblock {\em Proceedings of the 2nd International Conference on Learning
  Representations}.

\bibitem{miller2016variational}
Miller, A.~C., Foti, N.,  \& Adams, R.~P.
\newblock (2017) Variational boosting: {I}teratively refining posterior
  approximations.
\newblock {\em Proceedings of the 34th International Conference on Machine
  Learning} {\bf 70}, 2420--2429.

\bibitem{gershman2012nonparametric}
Gershman, S., Hoffman, M.,  \& Blei, D.
\newblock (2012) Nonparametric variational inference.
\newblock {\em Proceedings of the 29th International Coference on Machine
  Learning}.

\bibitem{kingma2014adam}
Kingma, D.~P. \& Ba, J.
\newblock (2014) Adam: {A} method for stochastic optimization.
\newblock {\em Proceedings of the 3rd International Conference on Learning
  Representations}.

\bibitem{hansen2003reducing}
Hansen, N., M{\"u}ller, S.~D.,  \& Koumoutsakos, P.
\newblock (2003) Reducing the time complexity of the derandomized evolution
  strategy with covariance matrix adaptation ({CMA-ES}).
\newblock {\em Evolutionary Computation} {\bf 11}, 1--18.

\bibitem{neal2003slice}
Neal, R.~M.
\newblock (2003) Slice sampling.
\newblock {\em Annals of Statistics} {\bf 31}, 705--741.

\bibitem{carpenter2017stan}
Carpenter, B., Gelman, A., Hoffman, M.~D., Lee, D., Goodrich, B., Betancourt,
  M., Brubaker, M., Guo, J., Li, P.,  \& Riddell, A.
\newblock (2017) Stan: {A} probabilistic programming language.
\newblock {\em Journal of Statistical Software} {\bf 76}.

\bibitem{gilks1994adaptive}
Gilks, W.~R., Roberts, G.~O.,  \& George, E.~I.
\newblock (1994) Adaptive direction sampling.
\newblock {\em The Statistician} {\bf 43}, 179--189.

\bibitem{kass1995bayes}
Kass, R.~E. \& Raftery, A.~E.
\newblock (1995) Bayes factors.
\newblock {\em Journal of the American Statistical Association} {\bf 90},
  773--795.

\bibitem{geyer1994estimating}
Geyer, C.~J.
\newblock (1994) Estimating normalizing constants and reweighting mixtures.
  ({T}echnical report).

\bibitem{knuth1992two}
Knuth, D.~E.
\newblock (1992) Two notes on notation.
\newblock {\em The American Mathematical Monthly} {\bf 99}, 403--422.

\bibitem{gelman2013bayesian}
Gelman, A., Carlin, J.~B., Stern, H.~S., Dunson, D.~B., Vehtari, A.,  \& Rubin,
  D.~B.
\newblock (2013) {\em {Bayesian} Data Analysis (3rd edition)}.
\newblock (CRC Press).

\bibitem{yao2018yes}
Yao, Y., Vehtari, A., Simpson, D.,  \& Gelman, A.
\newblock (2018) Yes, but did it work?: {E}valuating variational inference.
\newblock {\em arXiv preprint arXiv:1802.02538}.

\end{thebibliography}

}
}

\clearpage

\appendix
\part*{Supplementary Material}

\etocdepthtag.toc{mtappendix}
\etocsettagdepth{mtchapter}{none}
\etocsettagdepth{mtappendix}{subsubsection}

\setcounter{footnote}{0}
\setcounter{figure}{0}
\setcounter{table}{0}
\setcounter{equation}{0}
\renewcommand{\theequation}{S\arabic{equation}}
\renewcommand{\thetable}{S\arabic{table}}
\renewcommand{\thefigure}{S\arabic{figure}}

In this Supplement we include a number of derivations, implementation details, and additional results omitted from the main text.

Code used to generate the results in the paper is available at \url{https://github.com/lacerbi/infbench}. The VBMC algorithm is available at \url{https://github.com/lacerbi/vbmc}.

\tableofcontents

\section{Computing and optimizing the ELBO}
\label{sec:elbodetails}

For ease of reference, we recall the expression for the ELBO, for $\x \in \mathbb{R}^\nparams$,
\begin{equation}
\mathcal{F}\left[ q_{\qparams} \right] =  \mathbb{E}_{\qparams} \left[\log \frac{\like p(\x)}{q_{\qparams}(\x)} \right] = \mathbb{E}_{\qparams} \left[f(\x) \right] + \mathcal{H}[q_{\qparams}(\x)],
\end{equation}
with $\mathbb{E}_{\qparams} \equiv \mathbb{E}_{q_{\qparams}}$, and of the variational posterior,
\begin{equation} \label{eq:varpostapp}
q(\x) \equiv q_{\qparams}(\x) = \sum_{k = 1}^K w_k \normpdf{\x}{\mmu_k}{\sigma_k^2 \mathbf{\Sigma}},
\end{equation}
where $w_k$, $\mmu_k$, and $\sigma_k$ are, respectively, the mixture weight, mean, and scale of the $k$-th component, and $\mathbf{\Sigma} \equiv \text{diag}[{{\lambda}^{(1)}}^2,\ldots,{\lambda^{(\nparams)}}^2]$ is a diagonal covariance matrix common to all elements of the mixture.
The variational posterior for a given number of mixture components $K$ is parameterized by $\qparams \equiv (w_1,\ldots,w_K,\vmu_1, \ldots, \vmu_K, \sigma_1, \ldots, \sigma_K, \vlambda)$.

In the following paragraphs we derive expressions for the ELBO and for its gradient. Then, we explain how we optimize it with respect to the variational parameters.

\subsection{Stochastic approximation of the entropy}
\label{sec:entropy}

We approximate the entropy of the variational distribution via simple Monte Carlo sampling as follows. Let $\mat{R} = \text{diag}\left[ \bm{\lambda} \right]$ and $\ns$ be the number of samples per mixture component. We have
\begin{equation}
\begin{split}
\mathcal{H}\left[q(\x)\right] = & \, - \int q(\x) \log q(\x) d\x \\
\approx & \, - \frac{1}{\ns} \sum_{s = 1}^{\ns} \sum_{k = 1}^K w_k \log q(\sigma_k \R \eps_{s,k} + \mmu_k) \qquad \text{with} \quad \eps_{s,k} \sim \mathcal{N}\left(\bm{0}, \mathbb{I}_\nparams\right) \\
= & \, - \frac{1}{\ns} \sum_{s = 1}^{\ns} \sum_{k = 1}^K w_k \log q( \vxi_{s,k}) \qquad \text{with} \quad \vxi_{s,k} \equiv \sigma_k \R \eps_{s,k} + \mmu_k
\end{split}
\end{equation}
where we used the reparameterization trick separately for each component \cite{kingma2013auto,miller2016variational}. For VBMC, we set $\ns = 100$ during the variational optimization, and $\ns = 2^{15}$ for evaluating the ELBO with high precision at the end of each iteration.

\subsubsection{Gradient of the entropy}

The derivative of the entropy with respect to a variational parameter $\phi \in \{\mu, \sigma, \lambda \}$ (that is, not a mixture weight) is
\begin{equation}
\begin{split}
\frac{d}{d \phi} \mathcal{H}\left[q(\x)\right] \approx & \, - \frac{1}{\ns} \sum_{s = 1}^{\ns} \sum_{k = 1}^K w_k \frac{d}{d \phi} \log q(\vxi_{s,k}) \\
= & \, - \frac{1}{\ns} \sum_{s = 1}^{\ns} \sum_{k = 1}^K w_k \left(\frac{\partial}{\partial \phi} + \sum_{i=1}^\nparams \frac{d \xi^{(i)}_{s,k}}{d \phi} \frac{\partial}{\partial \xi^{(i)}_{s,k}} \right) \log q\left(\vxi_{s,k}\right) \\
= & \, - \frac{1}{\ns} \sum_{s = 1}^{\ns} \sum_{k = 1}^K \frac{w_k}{ q(\vxi_{s,k})} \sum_{i=1}^\nparams \frac{d \xi^{(i)}_{s,k}}{d \phi} \frac{\partial}{\partial \xi^{(i)}_{s,k}} \sum_{l = 1}^K w_l \normpdf{\vxi_{s,k}}{\mmu_l}{\sigma^2_l \bm{\Sigma}} \\
= & \, \frac{1}{\ns} \sum_{s = 1}^{\ns} \sum_{k = 1}^K \frac{w_k}{ q(\vxi_{s,k})} \sum_{i=1}^\nparams \frac{d \xi^{(i)}_{s,k}}{d \phi} \sum_{l = 1}^K w_l \frac{\xi_{s,k}^{(i)} - \mu_l^{(i)}}{ \left(\sigma_k {\lambda}^{(i)}\right)^2} \normpdf{\vxi_{s,k}}{\mmu_l}{\sigma^2_l \bm{\Sigma}} \\
\end{split}
\end{equation}
where from the second to the third row we used the fact that the expected value of the score is zero, $\mathbb{E}_{q(\vxi)}\left[\frac{\partial}{\partial \phi} \log q(\vxi)\right] = 0$.

In particular, for $\phi = \mu_j^{(m)}$, with $1 \le m \le \nparams$ and $1 \le j \le K$,
\begin{equation}
\begin{split}
\frac{d}{d \mu_j^{(m)}} \mathcal{H}\left[q(\x)\right] \approx & \, - \frac{1}{\ns} \sum_{s = 1}^{\ns} \sum_{k = 1}^K
\frac{w_k}{ q(\vxi_{s,k})}  \sum_{i=1}^\nparams \frac{d \xi^{(i)}_{s,k}}{d \mu_j^{(m)}} \frac{\partial}{\partial \xi^{(i)}_{s,k}}  \sum_{l = 1}^K w_l \normpdf{\vxi_{s,k}}{\mmu_l}{\sigma^2_l \bm{\Sigma}} \\
= & \, \frac{w_j}{\ns} \sum_{s = 1}^{\ns}
\frac{1}{ q(\vxi_{s,j})}  \sum_{l = 1}^K w_l \frac{\xi_{s,j}^{(m)} - \mu_l^{(m)}}{ \left(\sigma_l {\lambda}^{(m)}\right)^2} \normpdf{\vxi_{s,j}}{\mmu_l}{\sigma^2_l \bm{\Sigma}} \\
\end{split}
\end{equation}
where we used that fact that $\frac{d \xi^{(i)}_{s,k}}{d \mu_j^{(m)}} = \delta_{im} \delta_{jk}$.

For $\phi = \sigma_j$, with $1 \le j \le K$,
\begin{equation}
\begin{split}
\frac{d}{d \sigma_j} \mathcal{H}\left[q(\x)\right] \approx & \, - \frac{1}{\ns} \sum_{s = 1}^{\ns} \sum_{k = 1}^K
\frac{w_k}{ q(\vxi_{s,k})}  \sum_{i=1}^\nparams \frac{d \xi^{(i)}_{s,k}}{d \sigma_j} \frac{\partial}{\partial \xi^{(i)}_{s,k}}  \sum_{l = 1}^K w_l \normpdf{\vxi_{s,k}}{\mmu_l}{\sigma^2_l \bm{\Sigma}} \\
= & \, \frac{w_j}{K^2 \ns} \sum_{s = 1}^{\ns}
\frac{1}{ q(\vxi_{s,j})}  \sum_{i = 1}^\nparams \lambda^{(i)} \varepsilon_{s,j}^{(i)} \sum_{l = 1}^K w_l \frac{\xi_{s,j}^{(i)} - \mu_l^{(i)}}{ \left(\sigma_l {\lambda}^{(i)}\right)^2} \normpdf{\vxi_{s,j}}{\mmu_l}{\sigma^2_l \bm{\Sigma}} \\
\end{split}
\end{equation}
where we used that fact that $\frac{d \xi^{(i)}_{s,k}}{d \sigma_j} = \lambda^{(i)} \varepsilon_{s,j}^{(i)} \delta_{jk}$.

For $\phi = \lambda^{(m)}$, with $1 \le m \le \nparams$,
\begin{equation}
\begin{split}
\frac{d}{d \lambda^{(m)}} \mathcal{H}\left[q(\x)\right] \approx & \, - \frac{1}{\ns} \sum_{s = 1}^{\ns} \sum_{k = 1}^K
\frac{w_k}{ q(\vxi_{s,k})}  \sum_{i=1}^\nparams \frac{d \xi^{(i)}_{s,k}}{d \lambda^{(m)}} \frac{\partial}{\partial \xi^{(i)}_{s,k}}  \sum_{l = 1}^K w_l \normpdf{\vxi_{s,k}}{\mmu_l}{\sigma^2_l \bm{\Sigma}} \\
= & \, \frac{1}{\ns} \sum_{s = 1}^{\ns} \sum_{k = 1}^K
\frac{w_k \sigma_k \varepsilon_{s,k}^{(m)}}{ q(\vxi_{s,k})}  \sum_{l = 1}^K w_l \frac{\xi_{s,k}^{(m)} - \mu_l^{(m)}}{ \left(\sigma_l {\lambda}^{(m)}\right)^2}  \normpdf{\vxi_{s,k}}{\mmu_l}{\sigma^2_l \bm{\Sigma}} \\
\end{split}
\end{equation}
where we used that fact that $\frac{d \xi^{(i)}_{s,k}}{d \lambda^{(m)}} = \sigma_k \varepsilon_{s,k}^{(i)} \delta_{im}$.

Finally, the derivative with respect to variational mixture weight $w_j$, for $1 \le j \le K$, is
\begin{equation} \label{eq:gradentropy}
\begin{split}
\frac{\partial}{\partial w_j } \mathcal{H}\left[q(\x)\right] \approx & \;   \, - \frac{1}{\ns}\sum_{s = 1}^{\ns}  \left[ \log q(\vxi_{s,j}) + \sum_{k = 1}^K \frac{w_k}{q(\vxi_{s,k})} q_j(\vxi_{s,k}) \right].
 \end{split}
\end{equation}

\ignore{
\subsubsection{Moments and approximate KL-divergence}

The moments of a given variational approximation $q_\phi$ are
\begin{equation}
\begin{split}
\mathbb{E}[\x]_{q_\phi} \equiv & \; \vmu_{q_\phi} = \frac{1}{K} \sum_{k=1}^K \vmu_k, \\ 
\text{Cov}[\x]_{q_\phi} \equiv & \; \bm{\Sigma}_{q_\phi} = \left( \frac{1}{K} \sum_{k = 1}^K \sigma_k^2 \right) \bm{\Sigma} + \frac{1}{K} \sum_{k = 1}^K (\vmu_k - \vmu_{q_\phi}) (\vmu_k - \vmu_{q_\phi})^\top 
\end{split}
\end{equation}
which can be used to compute the approximate KL-divergence between two variational posteriors.
}

\subsection{Expected log joint}
\label{sec:logjoint}

For the expected log joint we have
\begin{equation} \label{eq:logjoint}
\begin{split}
\mathcal{G}[q(\x)] = \mathbb{E}_{\qparams}\left[f(\x)\right] = & \sum_{k = 1}^K  w_k \int \normpdf{\x}{\vmu_k}{\sigma_k^2 \bm{\Sigma}} f(\x) d \x \\
= &  \sum_{k = 1}^K w_k \mathcal{I}_k. \\
\end{split}
\end{equation}
To solve the integrals in Eq. \ref{eq:logjoint} we approximate $f(\x)$ with a Gaussian process (GP) with a \emph{squared exponential} (that is, rescaled Gaussian) covariance function,
\begin{equation} \label{eq:cov}
\mat{K}_{pq} = \kappa\left(\x_p, \x_q \right) = \sigma_f^2 \Lambda  \normpdf{\x_p}{\x_q}{\bm{\Sigma}_\ell} \qquad \text{with} \; \bm{\Sigma}_\ell = \text{diag}\left[{\ell^{(1)}}^2, \ldots, {\ell^{(\nparams)}}^2\right], 
\end{equation}
where $\Lambda \equiv \left(2\pi\right)^{\frac{\nparams}{2}} \prod_{i=1}^\nparams \ell^{(i)}$ is equal to the normalization factor of the Gaussian.\footnote{This choice of notation makes it easy to apply Gaussian identities used in Bayesian quadrature.}
For the GP we also assume a Gaussian likelihood with observation noise variance $\noise^2$ and, for the sake of exposition, a \emph{constant} mean function $m \in \mathbb{R}$. We will later consider the case of a \emph{negative quadratic} mean function, as per the main text.

\subsubsection{Posterior mean of the integral and its gradient}

The posterior predictive mean of the GP, given training data $\gpdata = \left\{\xx, \y\right\}$, where $\xx$ are $n$ training inputs with associated observed values $\y$, is
\begin{equation}
\overline{f}(\x) = \kappa(\x,\xx) \left[\kappa(\xx,\xx) + \noise^2 \mat{I}_n\right]^{-1} (\y - m) + m.
\end{equation}
Thus, for each integral in Eq. \ref{eq:logjoint} we have in expectation over the GP posterior
\begin{equation}
\begin{split}
\mathbb{E}_{f| \gpdata}\left[ \mathcal{I}_k\right]  = & \int \normpdf{\x}{\vmu_k}{\sigma_k^2 \bm{\Sigma}} \overline{f}(\x) d\x \\
= & \;  \left[ \sigma_f^2 \int \normpdf{\x}{\vmu_k}{\sigma_k^2 \bm{\Sigma}}  \normpdf{\x}{\xx}{\bm{\Sigma}_\ell}  d\x \right] \left[\kappa(\xx,\xx) + \noise^2 \mat{I}\right]^{-1} (\y - m) + m \\
= & \;  \z_k^\top \left[\kappa(\xx,\xx) + \noise^2 \mat{I}\right]^{-1} (\y - m) + m,
\end{split}
\end{equation}
where $\z_k$ is a $n$-dimensional vector with entries $z_k^{(p)} = \sigma_f^2 \normpdf{\vmu_k}{\x_p}{\sigma_k^2 \bm{\Sigma} + \bm{\Sigma}_\ell}$ for $1 \le p \le n$.
In particular, defining $\tau_k^{(i)} \equiv \sqrt{\sigma_k^2 {\lambda^{(i)}}^2 + {\ell^{(i)}}^2}$ for $1 \le i \le \nparams$,
\begin{equation}
z_k^{(p)} =  \frac{\sigma_f^2}{(2 \pi)^{\frac{\nparams}{2}} \prod_{i =1}^\nparams{\tau_k^{(i)}}}  \exp\left\{ -\frac{1}{2} \sum_{i=1}^\nparams \frac{\left(\mu_k^{(i)} - \x_p^{(i)}\right)^2}{ {\tau_k^{(i)}}^2} \right\}.
\end{equation}
We can compute derivatives with respect to the variational parameters $\phi \in (\mu, \sigma, \lambda)$ as
\begin{equation}
\begin{split}
\frac{\partial}{\partial \mu_j^{(l)} }z_k^{(p)} = & \; \delta_{jk}  \frac{\x_p^{(l)}-\mu_k^{(l)}}{{\tau_k^{(l)}}^2} z_k^{(p)}\\
\frac{\partial}{\partial \sigma_j }z_k^{(p)} = & \; \delta_{jk}   \sum_{i=1}^\nparams \frac{{\lambda^{(i)}}^2}{{\tau_k^{(i)}}^2} \left[  \frac{\left(\mu_k^{(i)}-\x_p^{(i)}\right)^2}{{\tau_k^{(i)}}^2} - 1 \right] \sigma_k z_k^{(p)} \\
\frac{\partial}{\partial \lambda^{(l)} }z_k^{(p)} = & \;  \frac{\sigma_k^2}{{\tau_k^{(l)}}^2} \left[  \frac{\left(\mu_k^{(l)}-\x_p^{(l)}\right)^2}{{\tau_k^{(l)}}^2} - 1 \right] {\lambda^{(l)}} z_k^{(p)} \\
\end{split}
\end{equation}
The derivative of Eq. \ref{eq:logjoint} with respect to mixture weight $w_k$ is simply $\mathcal{I}_k$.

\subsubsection{Posterior variance of the integral}

We compute the variance of Eq. \ref{eq:logjoint} under the GP approximation as \cite{ghahramani2003bayesian}
\begin{equation} \label{eq:varint}
\begin{split}
\text{Var}_{f|\X}[\mathcal{G}] = & \int \int q(\x) q(\x^\prime) C_{\gpdata}\left(f(\x),f(\x^\prime)\right)  \, d\x d\x^\prime \\
 = & \sum_{j = 1}^K \sum_{k = 1}^K w_j w_k \int \int \normpdf{\x}{\vmu_j}{\sigma_j^2 \bm{\Sigma}} \normpdf{\x^\prime}{\vmu_k}{\sigma_k^2 \bm{\Sigma}} C_{\gpdata}\left(f(\x),f(\x^\prime)\right) \, d\x d\x^\prime \\
 = & \sum_{j = 1}^K \sum_{k = 1}^K  w_j w_k  \mathcal{J}_{jk}
\end{split}
\end{equation}
where $C_{\gpdata}$ is the GP posterior predictive covariance,
\begin{equation}
C_{\gpdata}\left(f(\x),f(\x^\prime)\right) = \kappa(\x,\x^\prime) - \kappa(\x,\xx) \left[\kappa(\xx,\xx) + \noise^2 \mat{I}_n\right]^{-1} \kappa(\xx,\x^\prime).
\end{equation}
Thus, each term in Eq. \ref{eq:varint} can be written as
\begin{equation} \label{eq:varint1}
\begin{split}
\mathcal{J}_{jk} = & \int \int \normpdf{\x}{\vmu_j}{\sigma_j^2 \bm{\Sigma}} \left[ \sigma_f^2 \normpdf{\x}{\x^\prime}{\bm{\Sigma}_\ell} - \sigma_f^2 \normpdf{\x}{\xx}{\bm{\Sigma}_\ell} \left[\kappa(\xx,\xx) + \noise^2 \mat{I}_n\right]^{-1} \sigma_f^2 \normpdf{\xx}{\x^\prime}{\bm{\Sigma}_\ell} \right] \times \\
 & \quad \times \normpdf{\x^\prime}{\vmu_k}{\sigma_k^2 \bm{\Sigma}} \, d\x d\x^\prime \\
 =  & \, \sigma_f^2 \normpdf{\vmu_j}{\vmu_k}{\bm{\Sigma}_\ell + (\sigma_j^2 + \sigma_k^2)\bm{\Sigma}} - \z^\top_j \left[\kappa(\xx,\xx) + \noise^2 \mat{I}_n\right]^{-1} \z_k. \\
 \end{split}
\end{equation}

\ignore{
In particular, the self-interaction terms are
\begin{equation} \label{eq:varintself}
\begin{split}
\mathcal{J}_{kk} = &  \frac{\sigma_f^2}{(2\pi)^{\frac{\nparams}{2}} \prod_{i=1}^\nparams \tau_{kk}^{(i)}} - \z^\top_k \left[\kappa(\xx,\xx) + \noise^2 \mat{I}_n\right]^{-1} \z_k \qquad \text{ with } \; \tau_{kk}^{(i)} = \sqrt{2\sigma_k^2 {\lambda^{(i)}}^2 + {\ell^{(i)}}^2}\\
\end{split}
\end{equation}
for which the derivative with respect to variational parameter $\phi \in (\mu, \sigma, \lambda)$ is
\begin{equation} \label{eq:varintgrad}
\begin{split}
\frac{\partial}{\partial \phi} \mathcal{J}_{kk} = & - \frac{\sigma_f^2}{(2\pi)^{\frac{\nparams}{2}} \prod_{i=1}^\nparams {\tau_{kk}^{(i)}}^2} \frac{\partial}{\partial \phi} \left[\prod_{i=1}^\nparams {\tau_{kk}^{(i)}} \right] - 2 \z^\top_k \left[\kappa(\xx,\xx) + \noise^2 \mat{I}_n\right]^{-1} \frac{\partial \z_k}{\partial \phi}.
\end{split}
\end{equation}
The derivative of Eq. \ref{eq:varint} with respect to mixture weight $w_k$, considering only the self-interaction terms, is $2 w_k \mathcal{I}_k$.
}

\subsubsection{Negative quadratic mean function}

We consider now a GP with a \emph{negative quadratic} mean function,
\begin{equation}
m(\x) \equiv  m_\text{NQ}(\x) = m_0 - \frac{1}{2} \sum_{i=1}^\nparams \frac{\left(x^{(i)} - x_\text{m}^{(i)}\right)^2}{{\omega^{(i)}}^2}. 
\end{equation}
With this mean function, for each integral in Eq. \ref{eq:logjoint} we have in expectation over the GP posterior,
\begin{equation} \label{eq:ikquad}
\begin{split}
\mathbb{E}_{f| \gpdata}\left[ \mathcal{I}_k\right]  = & \;  \int\normpdf{\x}{\vmu_k}{\sigma_k^2 \bm{\Sigma}}  \left[  \sigma_f^2 \normpdf{\x}{\xx}{\bm{\Sigma}_\ell}   \left[\kappa(\xx,\xx) + \noise^2 \mat{I}\right]^{-1} (\y - m(\xx)) + m(\x) \right] d\x \\
= & \;  \z_k^\top \left[\kappa(\xx,\xx) + \noise^2 \mat{I}\right]^{-1} (\y - m(\xx)) + m_0 + \nu_k,
\end{split}
\end{equation}
where we defined
\begin{equation}
\begin{split}
\nu_k = & \; - \frac{1}{2} \sum_{i=1}^\nparams \frac{1}{{\omega^{(i)}}^2} \left({\mu_k^{(i)}}^2 + \sigma_k^2 {\lambda^{(i)}}^2 - 2 \mu_k^{(i)} x_\text{m}^{(i)} + {x_\text{m}^{(i)}}^2\right).
\end{split}
\end{equation}

\subsection{Optimization of the approximate ELBO}

In the following paragraphs we describe how we optimize the ELBO in each iteration of VBMC, so as to find the variational posterior that best approximates the current GP model of the posterior.

\subsubsection{Reparameterization}

For the purpose of the optimization, we reparameterize the variational parameters such that they are defined in a potentially unbounded space. The mixture means, $\vmu_k$, remain the same. We switch from mixture scale parameters $\sigma_k$ to their logarithms, $\log \sigma_k$, and similarly from coordinate length scales, $\lambda^{(i)}$, to $\log \lambda^{(i)}$. Finally, we parameterize mixture weights as unbounded variables, $\eta_k \in \mathbb{R}$, such that $w_k \equiv e^{\eta_k} / \sum_{l} e^{\eta_l}$ (softmax function). We compute the appropriate Jacobian for the change of variables and apply it to the gradients calculated in Sections \ref{sec:entropy} and \ref{sec:logjoint}.

\subsubsection{Choice of starting points}

In each iteration, we first perform a quick exploration of the ELBO landscape in the vicinity of the current variational posterior  by generating $n_\text{fast} \cdot K$ candidate starting points, obtained by randomly jittering, rescaling, and reweighting components of the current variational posterior. In this phase we also add new mixture components, if so requested by the algorithm, by randomly splitting and jittering existing components.
We evaluate the ELBO at each candidate starting point, and pick the point with the best ELBO as starting point for the subsequent optimization.

For most iterations we use $n_\text{fast} = 5$, except for the first iteration and the first iteration after the end of warm-up, for which we set $n_\text{fast} = 50$. 

\subsubsection{Stochastic gradient descent}

We optimize the (negative) ELBO via stochastic gradient descent, using a customized version of Adam \cite{kingma2014adam}. Our modified version of Adam includes a time-decaying learning rate, defined as
\begin{equation}
\alpha_t = \alpha_\text{min} + \left(\alpha_\text{max} - \alpha_\text{min}\right) \exp \left[ -\frac{t}{\tau} \right]
\end{equation}
where $t$ is the current iteration of the optimizer, $\alpha_\text{min}$ and $\alpha_\text{max}$ are, respectively, the minimum and maximum learning rate, and $\tau$ is the decay constant. We stop the optimization when the estimated change in function value or in the parameter vector across the past $n_\text{batch}$ iterations of the optimization goes below a given threshold. 

We set as hyperparameters of the optimizer $\beta_1 = 0.9$, $\beta_2 = 0.99$, $\epsilon \approx 1.49 \cdot 10^{-8}$ (square root of double precision), $\alpha_\text{min} = 0.001$, $\tau = 200$, $n_\text{batch} = 20$. We set $\alpha_\text{max} = 0.1$ during warm-up, and $\alpha_\text{max} = 0.01$ thereafter.

\section{Algorithmic details}

We report here several implementation details of the VBMC algorithm omitted from the main text.

\subsection{Regularization of acquisition functions}
\label{sec:regularization}

Active sampling in VBMC is performed by maximizing an acquisition function $a : \X \subseteq \mathbb{R}^\nparams \rightarrow [0,\infty)$, where $\X$ is the support of the target density. In the main text we describe two such functions, uncertainty sampling ($a_\text{us}$) and prospective uncertainty sampling ($a_\text{pro}$).

A well-known problem with GPs, in particular when using smooth kernels such as the squared exponential, is that they become numerically unstable when the training set contains points which are too close to each other, producing a ill-conditioned Gram matrix. Here we reduce the chance of this happening by introducing a correction factor as follows. For any acquisition function $a$, its regularized version $a^\text{reg}$ is defined as
\begin{equation} \label{eq:acqreg}
a^\text{reg}(\x) = a(\x) \exp \left\{ - \left(\frac{V^\text{reg}}{V_{\gpdata}(\x)} - 1 \right) \left|\left[V_{\gpdata}(\x) < V^\text{reg} \right]\right| \right\}
\end{equation}
where $V_{\gpdata}(\x)$ is the total posterior predictive variance of the GP at $\x$ for the given training set $\gpdata$, $V^\text{reg}$ a regularization parameter, and we denote with $|[\cdot]|$ \emph{Iverson's bracket} \cite{knuth1992two}, which takes value 1 if the expression inside the bracket is true, 0 otherwise. Eq. \ref{eq:acqreg} enforces that the regularized acquisition function does not pick points too close to points in $\gpdata$. For VBMC, we set $V^\text{reg} = 10^{-4}$.

\subsection{GP hyperparameters and priors}

The GP model in VBMC has $3\nparams+3$ hyperparameters, $\vgp = (\bm{\ell}, \sigma_f, \sigma_\text{obs}, m_0, \x_\text{m}, \bm{\omega})$. We define all scale hyperparameters, that is $\left\{\bm{\ell}, \sigma_f, \sigma_\text{obs}, \bm{\omega}\right\}$, in log space. 

We assume independent priors on each hyperparameter.
For some hyperparameters, we impose as prior a broad Student's $t$ distribution with a given mean $\mu$, scale $\sigma$, and $\nu = 3$ degrees of freedom. Following an empirical Bayes approach, mean and scale of the prior might depend on the current training set. For all other hyperparameters we assume a uniform flat prior.
 GP hyperparameters and their priors are reported in Table \ref{tab:gphyp}.

\begin{table}[ht]
\begin{tabular}{clcc}
Hyperparameter & Description & Prior mean $\mu$ & Prior scale $\sigma$ \\
\hline
$\log \ell^{(i)}$ & Input length scale ($i$-th dimension) & $\log$ SD$\left[\xx^{(i)}_\text{hpd}\right]$ & $\max \left\{ 2, \log \frac{\text{diam}\left[\xx^{(i)}_\text{hpd}\right] }{\text{SD}\left[\xx^{(i)}_\text{hpd}\right]} \right\} $ 
\\
$\log \sigma_f$ & Output scale & Uniform & --- \\
$\log \sigma_\text{obs}$ & Observation noise & $\log 0.001$ & 0.5 \\
$m_0$ & Mean function maximum & $\max \y_\text{hpd}$ & $\text{diam} \left[\y_\text{hpd}\right]$ \\
$x_\text{m}^{(i)}$ & Mean function location ($i$-th dim.) & Uniform & --- \\
$\log \omega^{(i)}$ & Mean function length scale ($i$-th dim.) & Uniform & --- \\
\hline
\end{tabular}
  \centering
  \vspace{.5em}
\caption{GP hyperparameters and their priors. See text for more information.}
\label{tab:gphyp}
\end{table}

In Table \ref{tab:gphyp}, SD$[\cdot]$ denotes the sample standard deviation and $\text{diam}[\cdot]$ the \emph{diameter} of a set, that is the maximum element minus the minimum. We define the \emph{high posterior density} training set, $\gpdata_\text{hpd} = \left\{\xx_\text{hpd}, \y_\text{hpd}\right\}$, constructed by keeping a fraction $f_\text{hpd}$ of the training points with highest target density values. For VBMC, we use $f_\text{hpd} = 0.8$ (that is, we only ignore a small fraction of the points in the training set).

\subsection{Transformation of variables}
\label{sec:changevars}

In VBMC, the problem coordinates are defined in an unbounded internal working space, $\x \in \mathbb{R}^\nparams$. All original problem coordinates $x_\text{orig}^{(i)}$ for $1\le i \le \nparams$ are independently transformed by a mapping $g_i: \X_\text{orig}^{(i)} \rightarrow \mathbb{R}$ defined as follows.

Unbounded coordinates are `standardized' with respect to the plausible box, $g_\text{unb}(x_\text{orig}) = \frac{x_\text{orig} - (\plb + \pub)/2}{\pub - \plb}$, where $\plb$ and $\pub$ are here, respectively, the plausible lower bound and plausible upper bound of the coordinate under consideration.

Bounded coordinates are first mapped to an unbounded space via a logit transform, $g_\text{bnd}(x_\text{orig}) = \log\left(\frac{z}{1-z}\right)$ with $z = \frac{x_\text{orig} - \lb}{\ub - \lb}$, where $\lb$ and $\ub$ are here, respectively, the lower and upper bound of the coordinate under consideration. The remapped variables are then `standardized' as above, using the remapped PLB and PUB values after the logit transform.

Note that probability densities are transformed under a change of coordinates by a multiplicative factor equal to the inverse of the determinant of the Jacobian of the transformation. Thus, the value of the observed log joint $y$ used by VBMC relates to the value $y_\text{orig}$ of the log joint density, observed in the original (untransformed) coordinates, as follows,
\begin{equation}
y(\x) = y^\text{orig}(\x_\text{orig}) - \sum_{i = 1}^\nparams \log g_i^\prime(\x_\text{orig}),
\end{equation}
where $g_i^\prime$ is the derivative of the transformation for the $i$-th coordinate, and $\x = g(\x_\text{orig})$. See for example \cite{carpenter2017stan} for more information on transformations of variables.

\subsection{Termination criteria}

The VBMC algorithm terminates when reaching a maximum number of target density evaluations, or when achieving long-term stability of the variational solution, as described below.

\subsubsection{Reliability index}

At the end of each iteration $t$ of the VBMC algorithm, we compute a set of reliability features of the current variational solution.
\begin{enumerate}
\item The absolute change in mean ELBO from the previous iteration: 
\begin{equation}
\rho_1(t) = \frac{\left|\mathbb{E}\left[\text{ELBO}(t)\right] - \mathbb{E}\left[\text{ELBO}(t-1)\right]\right|}{\Delta_\text{SD}},
\end{equation}
where $\Delta_\text{SD} > 0$ is a tolerance parameter on the error of the ELBO.
\item The uncertainty of the current ELBO: 
\begin{equation}
\rho_2(t) = \frac{\sqrt{\mathbb{V}\left[\text{ELBO}(t)\right]}}{\Delta_\text{SD}}.
\end{equation}
\item The change in symmetrized KL divergence between the current variational posterior $q_t \equiv q_{\qparams_t}(\x)$ and the one from the previous iteration:
\begin{equation} \label{eq:kldiff}
\rho_3(t) = \frac{\text{KL}(q_t || q_{t-1}) + \text{KL}(q_{t-1} || q_{t})}{2\Delta_\text{KL}},
\end{equation}
where for Eq. \ref{eq:kldiff} we use the Gaussianized KL divergence (that is, we compare solutions only based on their mean and covariance), and $\Delta_\text{KL} > 0$ is a tolerance parameter for differences in variational posterior.
\end{enumerate}
The parameters $\Delta_\text{SD}$ and $\Delta_\text{KL}$ are chosen such that $\rho_j \lesssim 1$, with $j = 1,2,3$, for features that are deemed indicative of a good solution.
For VBMC, we set $\Delta_\text{SD} = 0.1$ and $\Delta_\text{KL} = 0.01 \cdot\sqrt{\nparams}$.

The \emph{reliability index} $\rho(t)$ at iteration $t$ is obtained by averaging the individual reliability features $\rho_j(t)$.

\subsubsection{Long-term stability termination condition}

The long-term stability termination condition is reached at iteration $t$ when:
\begin{enumerate}
\item all reliability features $\rho_j(t)$ are below 1;
\item the reliability index $\rho$ has remained below 1 for the past $n_\text{stable}$ iterations (with the exception of at most one iteration, excluding the current one);
\item the slope of the ELCBO computed across the past $n_\text{stable}$ iterations is below a given threshold $\Delta_\text{IMPRO} > 0$, suggesting that the ELCBO is stationary.
\end{enumerate}
For VBMC, we set by default $n_\text{stable} = 8$ and $\Delta_\text{IMPRO} = 0.01$. For computing the ELCBO we use $\beta_\text{LCB} = 3$ (see Eq. 8 in the main text).

\subsubsection{Validation of VBMC solutions}

Long-term stability of the variational solution is \emph{suggestive} of convergence of the algorithm to a (local) optimum, but it should not be taken as a conclusive result without further validation.
In fact, without additional information, there is no way to know whether the algorithm has converged to a good solution, let alone to the global optimum. For this reason, we recommend to run the algorithm multiple times and compare the solutions, and to perform posterior predictive checks \cite{gelman2013bayesian}. See also \cite{yao2018yes} for a discussion of methods to validate the results of variational inference.



\section{Experimental details and additional results}

\subsection{Synthetic likelihoods}

\begin{figure}[htb]
  \includegraphics[width=\linewidth]{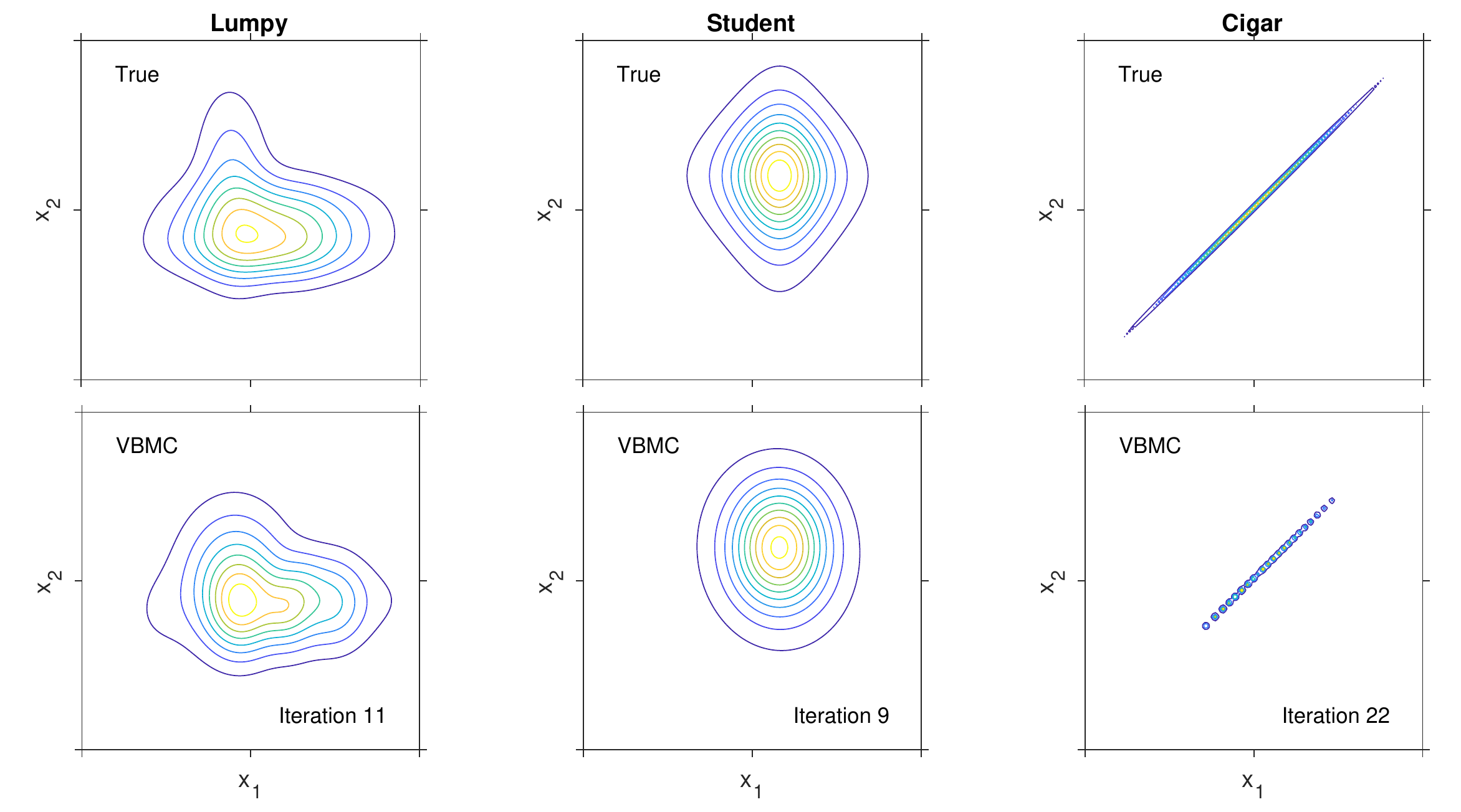}
\vspace{-1.75em}
  \caption{{\bf Synthetic target densities and example solutions.} \emph{Top:} Contour plots of two-dimensional synthetic target densities. \emph{Bottom:} Contour plots of example variational posteriors returned by VBMC, and iterations until convergence.
}
  \label{fig:synplot}
\end{figure}

We plot in Fig. \ref{fig:synplot} synthetic target densities belonging to the test families described in the main text (\emph{lumpy}, \emph{Student}, \emph{cigar}), for the $\nparams = 2$  case. We also plot examples of solutions returned by VBMC after reaching long-term stability, and indicate the number of iterations.

Note that VBMC, despite being overall the best-performing algorithm on the \emph{cigar} family in higher dimensions, still underestimates the variance along the major axis of the distribution. This is because the variational mixture components have axis-aligned (diagonal) covariances, and thus many mixture components are needed to approximate non-axis aligned densities. Future work should investigate alternative representations of the variational posterior to increase the expressive power of VBMC, while keeping its computational efficiency and stability.

\begin{figure}[htb!]
  \includegraphics[width=\linewidth]{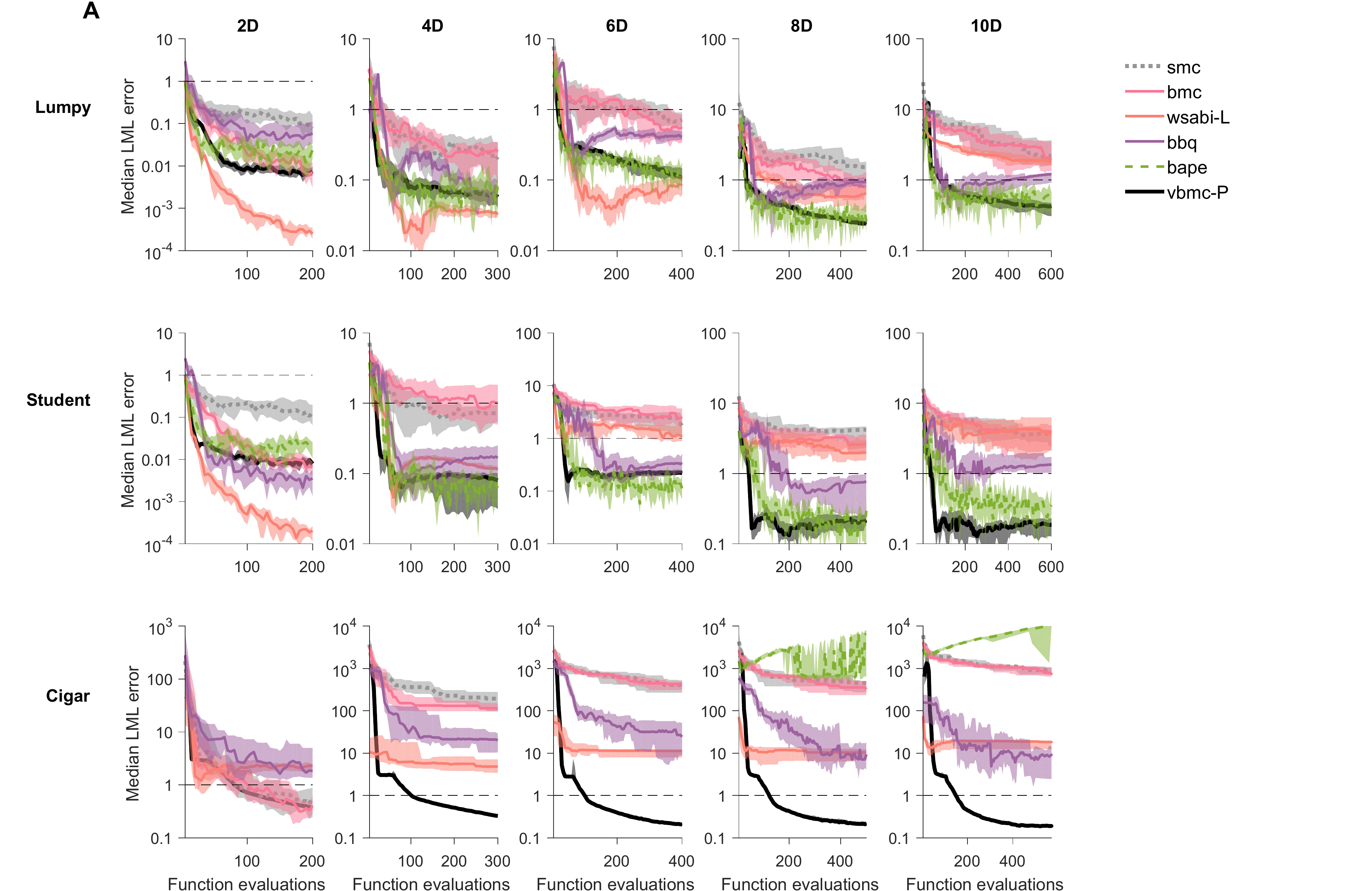}  
  \includegraphics[width=\linewidth]{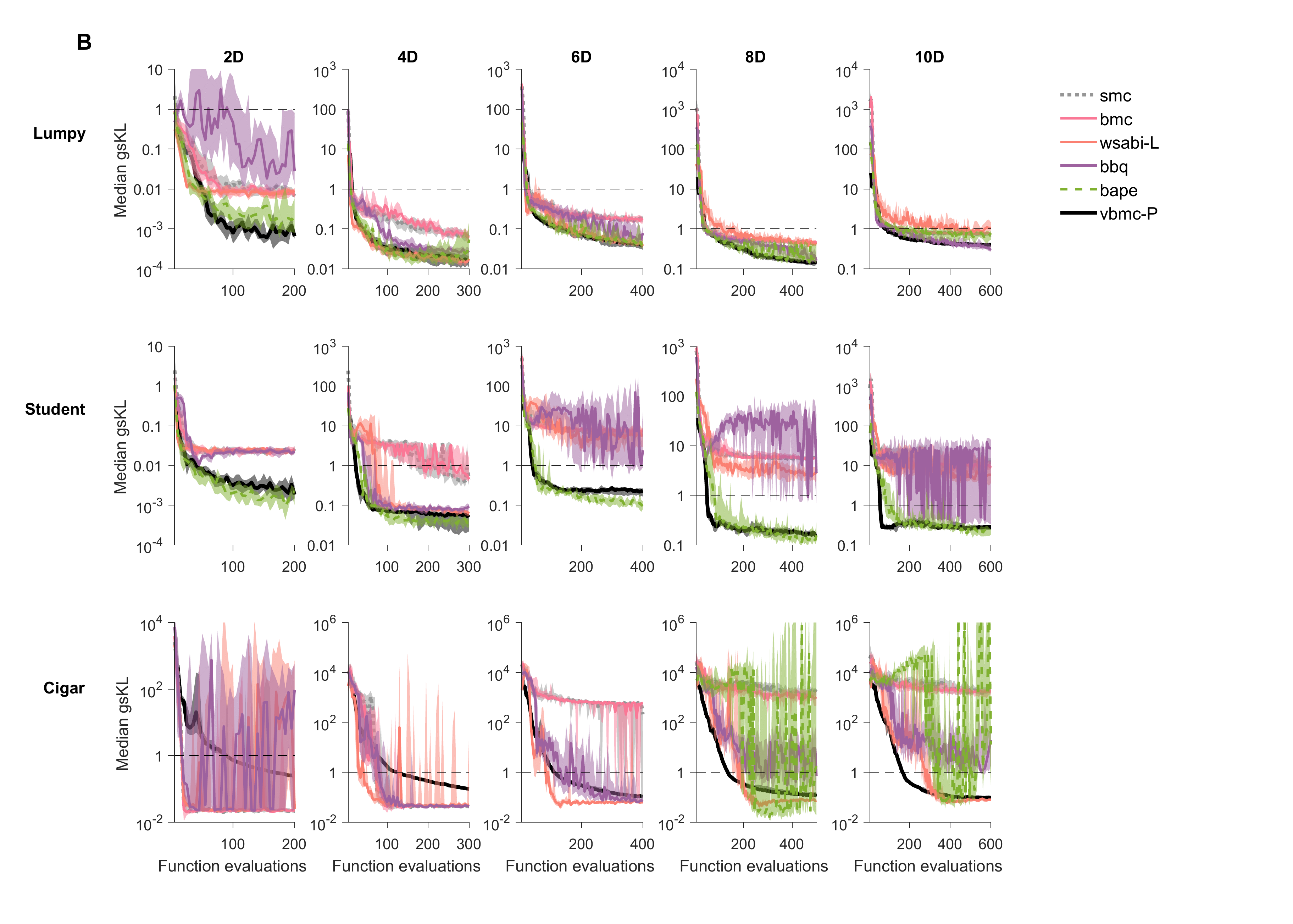}  
\vspace{-1.65em}
\centering
  \caption{{\bf Full results on synthetic likelihoods.} \textbf{A.} Median absolute error of the LML estimate with respect to ground truth, as a function of number of likelihood evaluations, on the \emph{lumpy} (top), \emph{Student} (middle), and \emph{cigar} (bottom) problems, for $\nparams \in \{2, 4, 6, 8, 10\}$ (columns). \textbf{B.} Median ``Gaussianized'' symmetrized KL divergence between the algorithm's posterior and ground truth.
For both metrics, shaded areas are 95 \% CI of the median,
and we consider a desirable threshold to be below one (dashed line). This figure reproduces Fig. 2 in the main text with more details. Note that panels here may have different vertical axes.
}
  \label{fig:syntheticfull}
\end{figure}

We plot in Fig. \ref{fig:syntheticfull} the performance of selected algorithms on the synthetic test functions, for $\nparams \in \{2,4,6,8,10 \}$. These results are the same as those reported in Fig. 2 in the main text, but with higher resolution. To avoid clutter, we exclude algorithms with particularly poor performance or whose plots are redundant with others. In particular, the performance of VBMC-U is virtually identical to VBMC-P here, so we only report the latter. Analogously, with a few minor exceptions, WSABI-M performs similarly or worse than WSABI-L across all problems. AIS suffers from the lack of problem-specific tuning, performing no better than SMC here, and the AGP algorithm diverges on most problems. Finally, we did not manage to get BAPE to run on the \emph{cigar} family, for $\nparams \le 6$, without systematically incurring in numerical issues with the GP approximation (with and without regularization of the BAPE acquisition function, as per Section \ref{sec:regularization}), so these plots are missing.

\subsection{Neuronal model}

As a real model-fitting problem, we considered in the main text a neuronal model that combines effects of filtering, suppression, and response nonlinearity, applied to two real data sets (one V1 and one V2 neurons) \cite{goris2015origin}. The purpose of the original study was to explore the origins of diversity of neuronal orientation selectivity in visual cortex via a combination of novel stimuli
(orientation mixtures) and modeling \cite{goris2015origin}. This model was also previously considered as a case study for a benchmark of Bayesian optimization and other black-box optimization algorithms \cite{acerbi2017practical}.

\subsubsection{Model parameters}

In total, the original model has 12 free parameters: 5 parameters specifying properties of a linear filtering mechanism, 2 parameters specifying nonlinear transformation of the filter output, and 5 parameters controlling response range and amplitude. For the analysis in the main text, we considered a subset of $\nparams = 7$ parameters deemed `most interesting' by the authors of the original study \cite{goris2015origin}, while fixing the others to their MAP values found by our previous optimization benchmark \cite{acerbi2017practical}.

The seven model parameters of interest from the original model, their ranges, and the chosen plausible bounds are reported in Table \ref{tab:gorisparams}.

\begin{table}[ht]
\begin{tabular}{clcccc}
Parameter & Description & $\lb$ & $\ub$ & $\plb$ & $\pub$ \\
\hline
$x_1$ & Preferred direction of motion (deg) & 0 & 360 & 90 & 270 \\
$x_2$ & Preferred spatial frequency (cycles per deg) & 0.05 & 15 & 0.5 & 10 \\
$x_3$ & Aspect ratio of 2-D Gaussian & 0.1 & 3.5 & 0.3 & 3.2 \\
$x_4$ & Derivative order in space & 0.1 & 3.5 & 0.3 & 3.2 \\
$x_5$ & Gain inhibitory channel & -1 & 1 & -0.3 & 0.3 \\
$x_6$ & Response exponent & 1 & 6.5 & 2 & 5 \\
$x_7$ & Variance of response gain & 0.001 & 10 & 0.01 & 1 \\
\hline
\end{tabular}
  \centering
  \vspace{.5em}
\caption{Parameters and bounds of the neuronal model (before remapping).}
\label{tab:gorisparams}
\end{table}

Since all original parameters are bounded, for the purpose of our analysis we remapped them to an unbounded space via a shifted and rescaled logit transform, correcting the value of the log posterior with the log Jacobian (see Section \ref{sec:changevars}). For each parameter, we set independent Gaussian priors in the transformed space with mean equal to the average of the transformed values of $\plb$ and $\pub$ (see Table \ref{tab:gorisparams}), and with standard deviation equal to half the plausible range in the transformed space.

\subsubsection{True and approximate posteriors}

\begin{figure}[htb!]
  \includegraphics[width=0.85\linewidth]{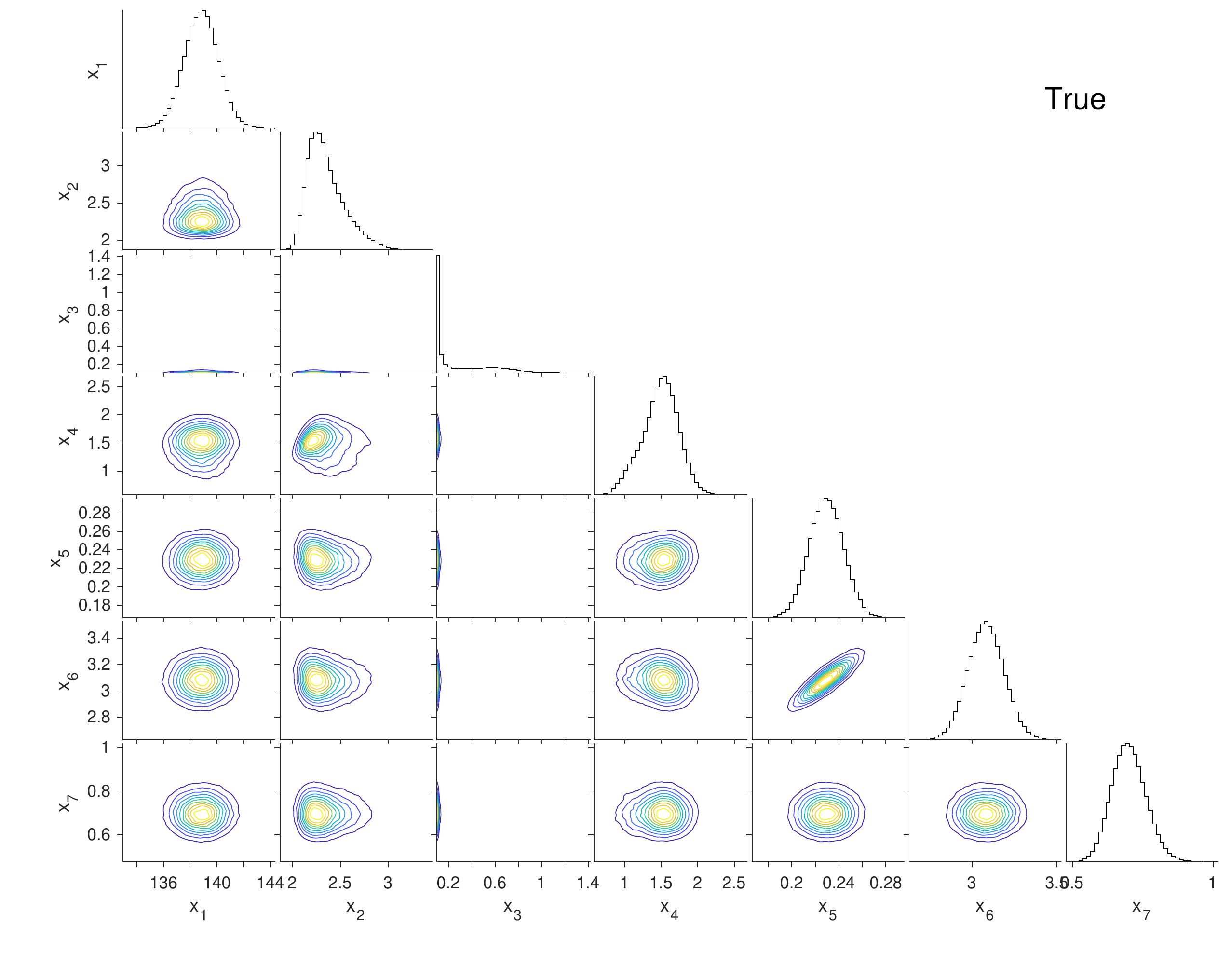}  
  \includegraphics[width=0.85\linewidth]{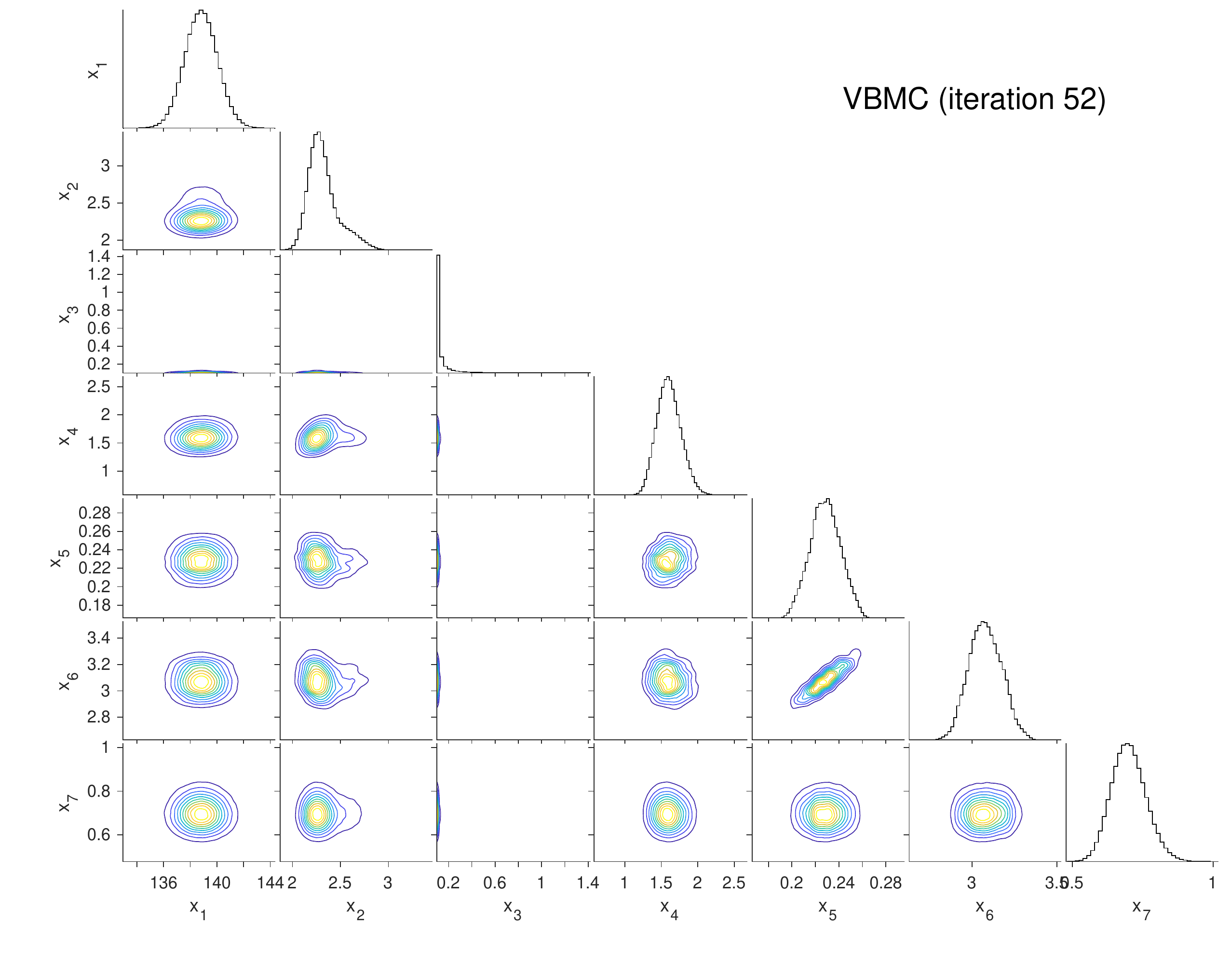}  
  \centering
\vspace{-1.75em}
  \caption{{\bf True and approximate posterior of neuronal model (V2 neuron).} \emph{Top}: Triangle plot of the `true' posterior (obtained via MCMC) for the neuronal model applied to the V2 neuron dataset. Each panel below the diagonal is the contour plot of the 2-D marginal distribution for a given parameter pair. Panels on the diagonal are histograms of the 1-D marginal distribution of the posterior for each parameter. \emph{Bottom}: Triangle plot of a typical variational solution returned by VBMC.
}
  \label{fig:goris2015post}
\end{figure}

We plot in Fig. \ref{fig:goris2015post} the `true' posterior obtained via extensive MCMC sampling for one of the two datasets (V2 neuron), and we compare it with an example variational solution returned by VBMC after reaching long-term stability (here in 52 iterations, which correspond to 260 target density evaluations).

We note that VBMC obtains a good approximation of the true posterior, which captures several features of potential interest, such as the correlation between the inhibition gain ($x_5$) and response exponent ($x_6$), and the skew in the preferred spatial frequency ($x_2$). The variational posterior, however, misses some details, such as the long tail of the aspect ratio ($x_3$), which is considerably thinner in the approximation than in the true posterior.

\clearpage
\section{Analysis of VBMC}

In this section we report additional analyses of the VBMC algorithm.

\subsection{Variability between VBMC runs}
\label{sec:robustness}

\begin{figure}[htb!]
  \includegraphics[width=\linewidth]{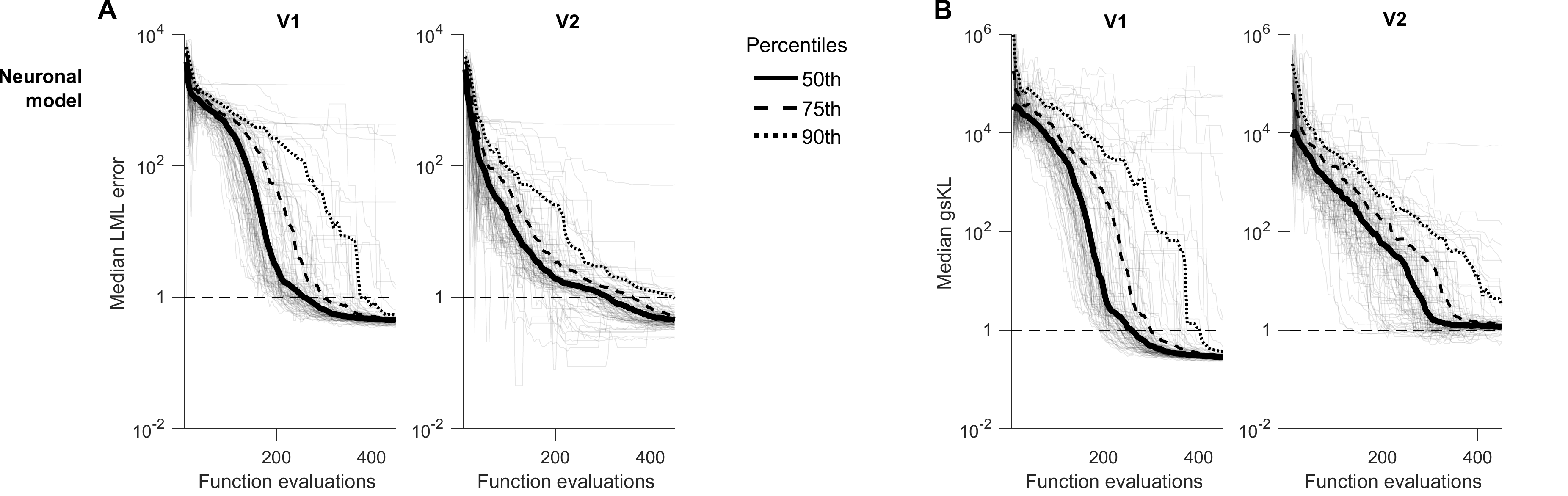}  
\vspace{-1.75em}
  \caption{{\bf Variability of VBMC performance.} \textbf{A.} Absolute error of the LML estimate, as a function of number of likelihood evaluations, for the two neuronal datasets. Each grey line is one of 100 distinct runs of VBMC. Thicker lines correspond to the 50th (median), 75th, and 90th percentile across runs (the median is the same as in Fig. 3 in the main text). 
 \textbf{B.} ``Gaussianized'' symmetrized KL divergence between the algorithm's posterior and ground truth, for 100 distinct runs of VBMC. See also Fig. 3 in the main text.
 }
  \label{fig:goris2015robust}
\end{figure}

In the main text we have shown the median performance of VBMC, but a crucial question for a practical application of the algorithm is the amount of variability between runs, due to stochasticity in the algorithm and choice of starting point (in this work, drawn uniformly randomly inside the plausible box).
We plot in Fig. \ref{fig:goris2015robust} the performance of one hundred runs of VBMC on the neuronal model datasets, together with the 50th (the median), 75th, and 90th percentiles. The performance of VBMC on this real problem is fairly robust, in that some runs take longer but the majority of them converges to quantitatively similar solutions.

\subsection{Computational cost}

The computational cost of VBMC stems in each iteration of the algorithm primarily from three sources: active sampling, GP training, and variational optimization. 
Active sampling requires repeated computation of the acquisition function (for its optimization), whose cost is dominated by calculation of the posterior predictive variance of the GP, which scales as $O(n^2)$, where $n$ is the number of training points. GP training scales as $O(n^3)$, due to inversion of the Gram matrix. Finally, variational optimization scales as $O(K n)$, where $K$ is the number of mixture components.
In practice, we found in many cases that in early iterations the costs are equally divided between the three phases, but later on both GP training and variational optimization dominate the algorithmic cost. In particular, the number of components $K$ has a large impact on the effective cost.

\begin{figure}[htb!]
  \includegraphics[width=0.7\linewidth]{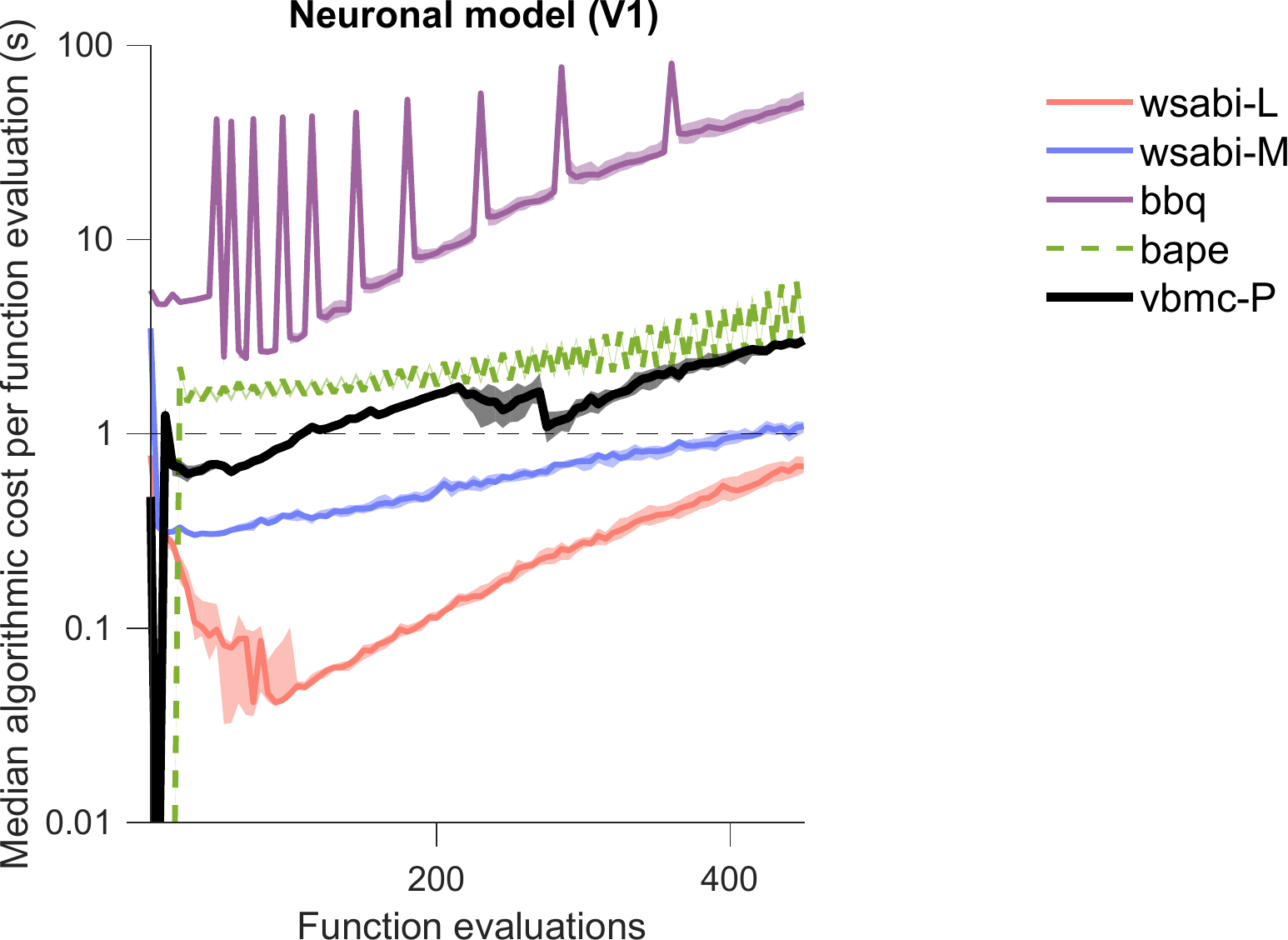}  
\centering
  \caption{{\bf Algorithmic cost per function evaluation.} Median algorithmic cost per function evaluation, as a function of number of likelihood function evaluations, for different algorithms performing inference over the V1 neuronal dataset. Shaded areas are 95 \% CI of the median.
}
  \label{fig:goris2015cost}
\end{figure}

As an example, we plot in Fig. \ref{fig:goris2015cost} the algorithmic cost per function evaluation of different inference algorithms that have been run on the V1 neuronal dataset (algorithmic costs are similar for the V2 dataset). We consider only methods which use active sampling with a reasonable performance on at least some of the problems. We define as algorithmic cost the time spent inside the algorithm, ignoring the time used to evaluate the log likelihood function. For comparison, evaluation of the log likelihood of this problem takes about 1 s on the reference laptop computer we used.
Note that for the WSABI and BBQ algoritms, the algorithmic cost reported here does not include the additional computational cost of sampling an approximate distrbution from the GP posterior (WSABI and BBQ, per se, only compute an approximation of the marginal likelihood).

VBMC on this problem exhibits a moderate cost of 2-3 s per function evaluation, when averaged across the entire run. Moreover, many runs would converge within 250-300 function evaluations, as shown in Figure \ref{fig:goris2015robust}, further lowering the effective cost per function evaluation.
For the considered budget of function evaluations, WSABI (in particular, WSABI-L) is up to one order of magnitude faster than VBMC. This speed is remarkable, although it does not offset the limited performance of the algorithm on more complex problems. WSABI-M is generally more expensive than WSABI-L (even though still quite fast), with a similar or slightly worse performance.
Here our implementation of BAPE results to be slightly more expensive than VBMC. Perhaps it is possible to obtain faster implementations of BAPE, but, even so, the quality of solutions would still not match that of VBMC (also, note the general instability of the algorithm).
Finally, we see that BBQ incurs in a massive algorithmic cost due to the complex GP approximation and expensive acquisition function used. Notably, the solutions obtained by BBQ in our problem sets are relatively good compared to the other algorithms, but still substantially worse than VBMC on all but the easiest problems, despite the much larger computational overhead.

The dip in cost that we observe in VBMC at around 275 function evaluations is due to the switch from GP hyperparameter sampling to optimization.
The cost of BAPE oscillates because of the cost of retraining the GP model and MCMC sampling from the approximate posterior every 10 function evaluations. Similarly, by default BBQ retrains the GP model ten times, logarithmically spaced across its run, which appears here as logarithmically-spaced spikes in the cost.

\subsection{Analysis of the samples produced by VBMC}
\label{sec:control}

We report the results of two control experiments to better understand the performance of VBMC.

\begin{figure}[htb]
  \includegraphics[width=\linewidth]{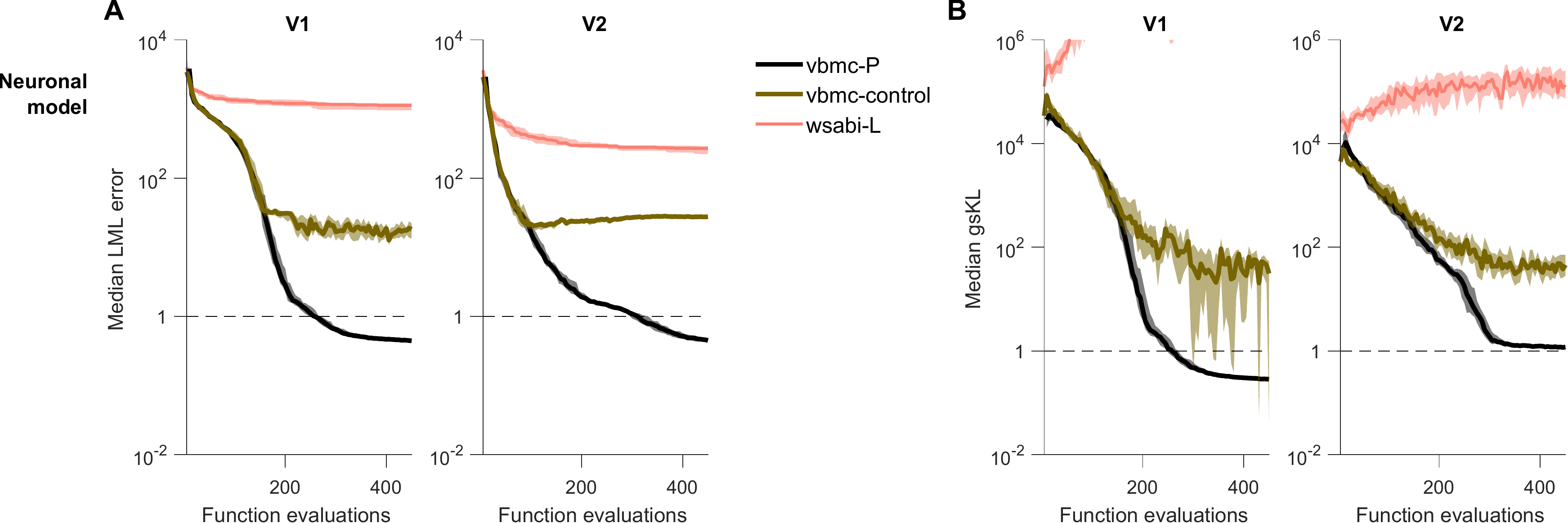}
\vspace{-1.75em}
  \caption{{\bf Control experiments on neuronal model likelihoods.} \textbf{A.} Median absolute error of the LML estimate, as a function of number of likelihood evaluations, for two distinct neurons ($\nparams = 7$). For the control experiment, here we computed the LML with WSABI-L trained on VBMC samples.
 \textbf{B.} Median ``Gaussianized'' symmetrized KL divergence between the algorithm's posterior and ground truth.
 For this control experiment, we produced posterior samples from a GP directly trained on the log joint at the samples produced by VBMC. 
 For both metrics, shaded areas are 95\% CI of the median, and we consider a desirable threshold to be below one (dashed line). See text for more details, and see also Fig. 3 in the main text.
}
  \label{fig:goris2015control}
\end{figure}

For the first control experiment, shown in Fig. \ref{fig:goris2015control}A, we estimate the log marginal likelihood (LML) using the WSABI-L approximation trained on samples obtained by VBMC (with the $a_\text{pro}$ acquisition function).
The LML error of WSABI-L trained on VBMC samples is lower than WSABI-L alone, showing that VBMC produces higher-quality samples and, given the same samples, a better approximation of the marginal likelihood. The fact that the LML error is still substantially higher in the control than with VBMC alone demonstrates that the error induced by the WSABI-L approximation can be quite large.

For the second control experiment, shown in Fig. \ref{fig:goris2015control}B, we produce $2 \cdot 10^4$ posterior samples from a GP directly trained on the log joint distribution at the samples produced by VBMC. 
The quality of this posterior approximation is better than the posterior obtained by other methods, although generally not as good as the variational approximation (in particular, it is much more variable).
While it is possible that the posterior approximation via direct GP fit could be improved, for example by using ad-hoc methods to increase the stability of the GP training procedure, this experiment shows that VBMC is able to reliably produce a high-quality variational posterior.

\end{document}